\documentclass[journal]{IEEEtran}

\usepackage{cite}
\usepackage{amsmath}
\usepackage{amssymb}
\usepackage{times}
\usepackage{multirow}
\usepackage{color}
\usepackage[switch]{lineno}
\usepackage{graphicx}
\usepackage{subfigure}
\usepackage{url}

\usepackage{algorithm}
\usepackage{algorithmic}
\DeclareMathOperator*{\argmax}{arg\,max}

\begin{document}
\title{Learning to Segment Object Candidates via Recursive Neural Networks}

\author{Tianshui~Chen, Liang~Lin, Xian~Wu, Nong~Xiao, and Xiaonan~Luo 
\thanks{Corresponding author is Xiaonan Luo.

T. Chen, L. Lin, X. Wu, and N. Xiao are with the School of Data and Computer Science, Sun Yat-sen University, Guangzhou 510006, China. (E-mail: tianshuichen@gmail.com, linliang@ieee.org, sysuwuxian@gmail.com, xiaon6@mail.sysu.edu.cn).  

X. Luo is with the School of Computer Science and Information Security, Guilin University of Electronic Technology, Guilin 541004, China (E-mail: luoxn@guet.edu.cn).
}}

\maketitle

\begin{abstract}
To avoid the exhaustive search over locations and scales, current state-of-the-art object detection systems usually involve a crucial component generating a batch of candidate object proposals from images. In this paper, we present a simple yet effective approach for segmenting object proposals via a deep architecture of recursive neural networks (ReNNs), which hierarchically groups regions for detecting object candidates over scales. Unlike traditional methods that mainly adopt fixed similarity measures for merging regions or finding object proposals, our approach adaptively learns the region merging similarity and the objectness measure during the process of hierarchical region grouping. Specifically, guided by a structured loss, the ReNN model jointly optimizes the cross-region similarity metric with the region merging process as well as the objectness prediction. During inference of the object proposal generation, we introduce randomness into the greedy search to cope with the ambiguity of grouping regions. Extensive experiments on standard benchmarks, e.g., PASCAL VOC and ImageNet, suggest that our approach is capable of producing object proposals with high recall while well preserving the object boundaries and outperforms other existing methods in both accuracy and efficiency.
\end{abstract}


\IEEEpeerreviewmaketitle

\section{Introduction}
\label{intro}
\begin{figure}[!t]
\centering
\includegraphics[width=1.0\linewidth]{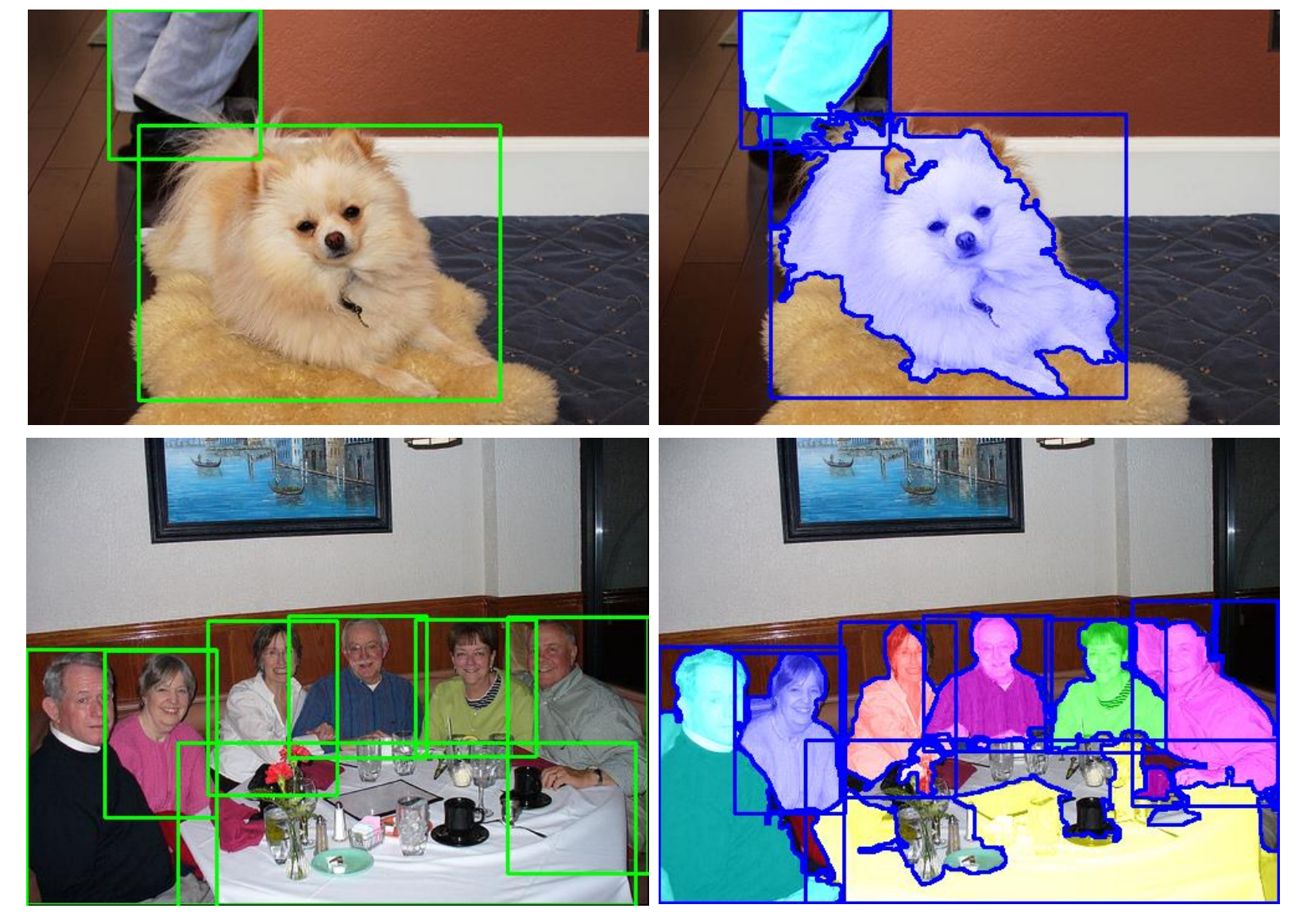}
\caption{Some object proposals (indicated by the blue boxes) generated by our approach. Our results match well with the ground-truth (indicated by the green boxes), and also preserve the object boundaries (indicated by the blue silhouettes inside the boxes).}
\label{fig:visualization_results1}
\end{figure}

\IEEEPARstart{O}{bject} proposal generation, which aims to identify a small set of region proposals where objects are likely to occur, benefits a wide range of applications such as generic object detection \cite{girshick2014rich,he2014spatial}, object recognition \cite{wei2016hcp,wang2017multi,chen2018recurrent} and object discovery \cite{lee2011learning,cho2015unsupervised}. Usually, a good object proposal method is desired to be capable of not only recalling all existing objects over scales and locations but also preserving their boundaries, for example in Figure \ref{fig:visualization_results1}.

The challenges of object proposal lie in the presence of severe occlusion, variations in object shapes, and the lack of category information. Most of the current methods ~\cite{uijlings2013selective,manen2013prime,xiao2015complexity} tackle these difficulties through bottom-up region grouping or segmentation. Those methods mainly involve two crucial components, i.e., cross-region similarity metric and region merging algorithm. The similarity metric is utilized to measure whether two adjacent regions should be merged, and the merging algorithm performs the inference process that groups pairs of regions into super-regions and finally generates object proposals. Thus, object proposal generation methods based on region grouping basically follow the pipeline:
they assign a higher similarity score to the adjacent regions if it is confident that the regions belong to the same class, and recursively merge the adjacent regions with highest score. Despite of acknowledged successes, these approaches usually require elaborative tuning or setting (e.g., manually designed cross-region similarity metric), limiting their performance in complex environments.

In this work, we develop a novel hierarchical region grouping approach for generating and segmenting object proposals by learning a recursive neural network (ReNN). In our ReNN architecture, we incorporate the cross-region similarity metric learning into bottom-up region merging process for end-to-end training. In particular, we define a structured loss that penalizes the incorrect merging candidates by measuring the similarity of adjacent regions and the objectness. In this way, our model explicitly optimizes the cross-region similarity learning and objectness prediction within the recursive iterations. Interestingly, the forward process of ReNN finely accords with the traditional bottom-up region grouping pipeline, leading to a very natural embedding of the two crucial components (i.e., cross-region similarity metric and merging algorithm). Moreover, the objectness score is also learned with the ReNN training, bringing the benefit of fast rejecting the false positive samples.

Obviously, the greedy merging algorithms, that recursively merge two regions with highest merging scores, can be applied for inference with the ReNN model ~\cite{socher2011parsing}. However, the performance of greedy methods depends heavily on the accuracy of merging scores, since greedy merging is generally sensitive to noise or local minima. In the task of object proposal generation, once a segment of an object is incorrectly merged with the background or other objects, this object has little possibility to be recalled. In addition, we experimentally found that greedy merging leads to incorrect object proposals easily, especially when one segment of an object has similar appearance with background or other surrounding objects. 
To alleviate this issue, we propose a randomized merging algorithm that introduces randomness in the recursive inference procedure. Instead of merging a pair of neighbouring regions with highest similarity score, we search for $k$ pairs with top $k$ highest similarities, and then randomly pick one pair according to a distribution constructed by their scores. The process is repeated for $K$ times, thus that errors occurred at one random merging process can be corrected in other processes.
In this way, it can help to recall more incorrectly merged objects. Figure \ref{fig:visualization_results1} shows some examples of object proposals generated by our approach.

The key contribution of this work is a deep architecture of recursive neural networks for generating object proposals and preserving their boundaries. This framework jointly optimizes the cross-region similarity and objectness measure together with the hierarchical region grouping process, which is original in literature of object segmentation and detection. Moreover, we design a randomized region merging algorithm with the recursive neural network learning, which introduces randomness to handle the inherent ambiguities of composing regions into candidate objects and thus causes a notable gain in object recall rate. Extensive experimental evaluation and analysis on standard benchmarks (e.g., PASCAL VOC and ImageNet) are provided, demonstrating that our method achieves superior performances over existing approaches in both accuracy and efficiency.

The remainder of the paper is organized as follows. Section \ref{sec:related_works} presents a review of the related works. We then introduce our approach and optimization algorithm in detail in Section \ref{sec:deep_objectness_model} and Section \ref{sec:optimization}, respectively. Experimental results, comparisons and analysis are exhibited in Section \ref{sec:experiment}. Section \ref{sec:conclusion} concludes the paper.

\section{Related Work}
\label{sec:related_works}
Many efforts have been dedicated to object proposal generation. Here we roughly divide existing methods into two categories: top-down window-based scoring and bottom-up region grouping, according to their computation process.

\subsection{Window-based Scoring}
This category of methods ~\cite{ren2015faster,zitnick2014edge,cheng2014bing,alexe2010object,alexe2012measuring} attempt to distinguish object proposals directly from the surrounding background through assigning an objectness score to each candidate sub-window. The objectness measures are usually defined in diverse ways, and object proposals generated by sliding windows are then ranked and thresholded by their objectness scores. As a pioneer work, Alexe et al. ~\cite{alexe2010object} employed saliency cue to measure the objectness of a given window, which was further improved by ~\cite{zhang2011proposal} with learning methods and more complicated features. However, these methods may suffer from expensive computational cost, since they require to search over all locations and scales in images. Recently, to address this problem, BING ~\cite{cheng2014bing} and Edge Box ~\cite{zitnick2014edge} exploited very simple features such as gradient and contour information to score the windows, and achieved very high computational efficiency. Alternatively, Ren et al. ~\cite{ren2015faster} proposed a deep learning method based on fully convolutional networks (FCNs) ~\cite{long2014fully} to score windows over scales and locations efficiently. Nonetheless, this method may not locate object accurately, since experimental results show that the recall rate deteriorates as the Intersection over Union (IoU) threshold increases.

\subsection{Region Grouping}
This branch of researches ~\cite{uijlings2013selective,krahenbuhl2014geodesic,carreira2012cpmc,manen2013prime,arbelaez2014multiscale,rantalankila2014generating, bergh2013online} cast the object proposal generation as a process of hierarchical region segmentation or partition. Starting from an initial over-segmentation, these methods usually adopt a cross-region similarity / distance metric \cite{lin2017cross,chen2016deep} that works together with region merging algorithms. As a representative example of these methods, Uijlings et al. ~\cite{uijlings2013selective} leveraged four types of low-level features (e.g., color, texture etc.) for similarity computing and generated object proposals via hierarchical greedy grouping. Using similar features with ~\cite{uijlings2013selective}, Manen et al. ~\cite{manen2013prime} learned the merging probabilities and introduced a randomized prim algorithm for region grouping. Following similar hierarchical grouping methods,  Wang et al. \cite{wang2015object} proposed a multi-branch hierarchical segmentation method via learning multiple merging strategies at each step.  Arbel\'aez et al. ~\cite{arbelaez2014multiscale} constructed hierarchical segmentations and explored the combinatorial space to combine multi-scale regions into proposals. Xiao et al. \cite{xiao2015complexity} proposed a complexity-adaptive distance metric for grouping the neighbouring super-pixels. It combined a low-complexity distance and a high-complexity distance to adapt different complexity levels. Kr{\"a}henb{\"u}hl and Koltun \cite{krahenbuhl2014geodesic} trained classifiers to adaptively place seeds to hit the objects in the image, and identified a small set of level sets as object proposals for each seed. This method was further improved by ensembling multiple models to generate more diverse proposals \cite{krahenbuhl2015learning}. Rantalankila et al. ~\cite{rantalankila2014generating} integrated local region merging and global graph-cut to generate proposals. 
Due to their high localization accuracy, they are adopted in many state-of-the-art object detection \cite{girshick2014rich,he2014spatial} and object discovery \cite{cho2015unsupervised} algorithms. However, these mentioned methods mainly adopt fixed similarity measures for merging regions or finding object proposals, leading to suboptimal performances when handling complex cases. In contrast, our approach adaptively learns the region merging similarity and the objectness measure during the process of hierarchical region grouping. Moreover, our method also introduces randomness into the bottom-up searching of region composition and yields significant improvement over existing methods.

\begin{figure*}[!t]
\centering
\includegraphics[width=1.0\linewidth]{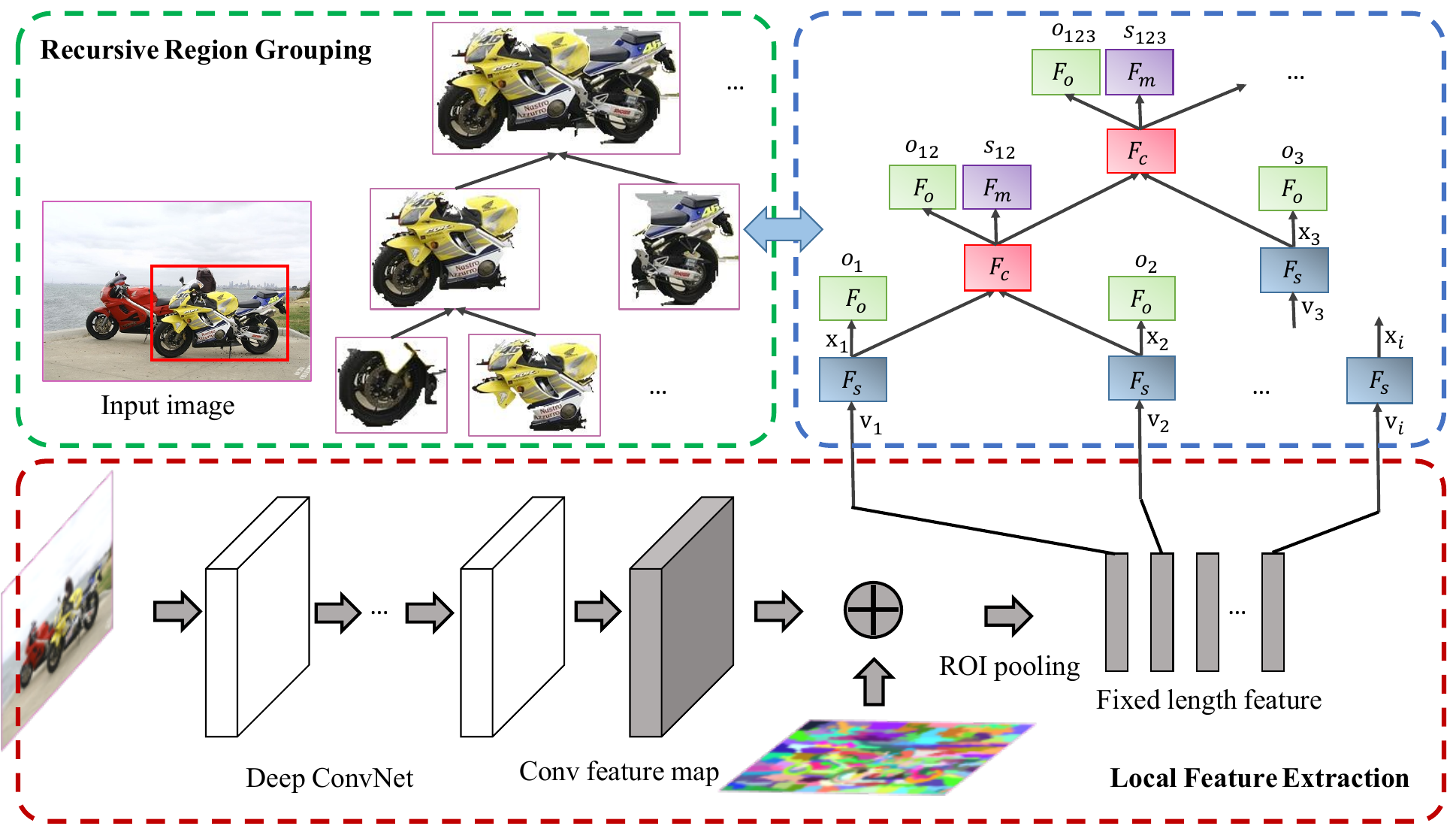}
\caption{An overview of our proposed object proposal segmentation framework. The bottom shows local feature extraction, and the top illustrates bottom-up recursive region grouping process. The four modules, $F_s$, $F_c$, $F_m$ and $F_o$, work cooperatively to group regions for generating object proposals.}
\label{fig:rnn}
\end{figure*}

\section{Framework of Segmenting Object Proposals}
\label{sec:deep_objectness_model}
In this section, we introduce our approach in detail. The input image is first over-segmented into $N$ regions with the efficient graph-based method ~\cite{felzenszwalb2004efficient}. The Fast R-CNN ~\cite{girshick2015fast} is used to extract local features for each region. We then design a recursive neural network to group regions and simultaneously predict the associated objectness scores for corresponding proposals. Furthermore,
we propose a randomized merging algorithm, which introduces randomness into recursive inference procedure to cope with the inherent ambiguities in the process of merging regions. Figure \ref{fig:rnn} gives an illustration of our proposed framework.

\subsection{Local Feature Extraction}

Since deep features have shown significant improvement than hand-crafted features on various vision tasks \cite{simonyan2014very,chen2016disc,lin2016deep,lin2018active,chen2018knowledge,wang2018deep}, we utilize the Fast RCNN ~\cite{girshick2015fast} architecture to extract deep local features for each region. The architecture consists of 16 convolutional layers, the same as VGG16-net ~\cite{simonyan2014very}, followed by the region of interests (ROI) pooling layer. Specifically, given an input image, our approach first over-segments it into $N$ regions with the efficient graph-based method ~\cite{felzenszwalb2004efficient} and obtains the box for each region that tightly bounds this region. To achieve a better trade-off between speed and accuracy, we follow \cite{girshick2015fast} to resize the input image, thus that the short side of the image is 600, remaining the aspect ratio unchanged. The sixteen convolutional layers take the resized image as input, and produce a pooling of corresponding size feature maps. The ROI pooling layer subsequently extracts a fixed length feature vector for each region.

\subsection{Recursive Neural Networks}
We first present some notations that would be used throughout this article. Let $\mathbf{v}_{i}$ denote the local features of the $i$-th region, and $\mathbf{x}_i$ denote the corresponding semantic features. $\sigma\left(\cdot\right)$ denotes the rectified linear unit (ReLU), where $\sigma\left(x\right)=\max(0,x)$.

\begin{figure}[htp]
\centering
\includegraphics[width=0.85\linewidth]{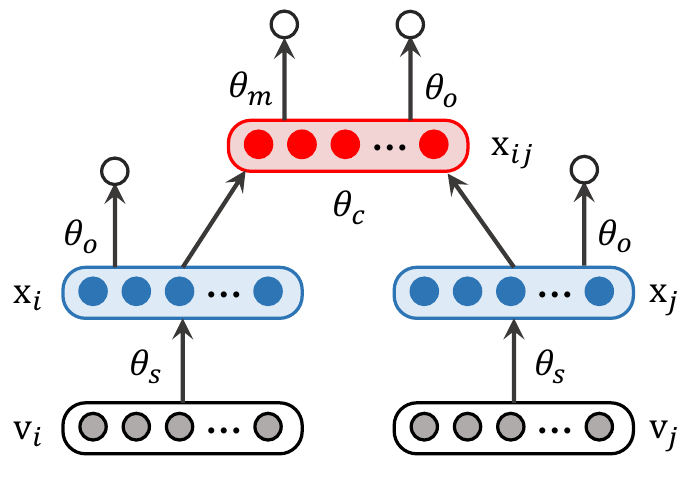}
\caption{Illustration of the recursive neural network in our proposed framework. This network computes the scores for merging decision and objectness scores of all regions.}
\label{fig:rnn_network}
\end{figure}

The core of this framework is the ReNN, which aims to group the regions and simultaneously predict the objectness scores for corresponding proposals in a recursive manner. The ReNN architecture is depicted in Figure \ref{fig:rnn_network}. The ReNN comprises four modules, i.e., semantic mapper, feature combiner, merging scorer and objectness scorer. Semantic mapper transforms the local features to semantic space which can be further propagated to their parent nodes. Feature combiner computes the joint semantic representations of all neighbouring child nodes. Given joint semantic representations, merging scorer calculates the score indicating the confidence that two nodes should be merged. Feature combiner merges the neighbouring nodes according to merging scores, and obtains a hierarchical tree structural segmentations, each of which corresponds to a candidate of proposal. Objectness scorer computes a score which estimates the likelihood of the candidate containing an object. These four modules work cooperatively for proposal segmentation, as illustrated in Figure \ref{fig:rnn}. We describe these four modules in the following.

\subsubsection{Semantic Mapper}
Semantic mapper $F_{s}$ is a simple feed-forward operator to map the local features into the semantic space in which the combiner operates on. It can be expressed as,
\begin{equation}
\centering
\begin{aligned}
\mathbf{x}_{i}=F_{s}\left(\mathbf{v}_i;\theta_{s}\right)=\sigma\left(W_{s}\mathbf{v}_i+\mathbf{b}_{s}\right),
\label{eqn:semantic_mapping}
\end{aligned}
\end{equation}
$F_{s}$ captures the region semantic representation, and propagates it to its parent regions through the tree hierarchical structure. To better balance the computational efficiency and accuracy, we empirically set the dimensionality of local features $\mathbf{v}_{i}$ as 18,432 ($6\times6\times512$), and that of semantic features $\mathbf{x}_{i}$ as 256. Hence, the semantic mapper is a one-layer fully-connected network, with 18,432 input and 256 output neurons, followed by the rectified linear unit. $\theta_{s}=\{W_s, \mathbf{b}_{s}\}$ are the learnt parameters, in which $W_s$ and $\mathbf{b}_{s}$ are the weight matrix and bias of the fully-connected layer, respectively.

\subsubsection{Feature combiner}
Feature combiner $F_{c}$ recursively takes the semantic features of its two child nodes as input, and maps them to the semantic features of the parent node, formulated as,
\begin{equation}
\begin{aligned}
\mathbf{x}_{i,j}=F_{c}\left([\mathbf{x}_i, \mathbf{x}_j];\theta_{c}\right)=\sigma\left(W_{c}[\mathbf{x}_i, \mathbf{x}_j]+\mathbf{b}_{c}\right),
\label{eqn:feature_combining}
\end{aligned}
\end{equation}
$F_{c}$ aggregates the semantic information of the two child nodes and obtains the semantic representation of the merged node. It takes semantic features of the original regions as leaf nodes, and recursively aggregates them to the root node in a bottom-up manner. In order to ensure the recursive procedure can be applied, the dimensionality of parent node features is set the same as that of child node features. Thus, the architecture of the feature combiner is identical to that of the semantic mapper, except that it has 512 ($2\times256$) input neurons. Similarly, $\theta_{c}=\{W_c, \mathbf{b}_{c}\}$ are its learnt parameters, where $W_c$ and $\mathbf{b}_{c}$ are the weight matrix and bias, respectively.

\subsubsection{Merging scorer}
Given the joint semantic features of two neighbouring nodes, merging scorer $F_{m}$ computes a score that indicates the confidence that whether two nodes should be merged, expressed as
\begin{equation}
\begin{aligned}
s_{i,j}=F_{m}\left(\mathbf{x}_{i,j};\theta_{m}\right)=W_{m}\mathbf{x}_{i,j}+\mathbf{b}_m,
\label{eqn:merge_scoring}
\end{aligned}
\end{equation}
The scores determine the pair that should be merged first in both learning and inference stages. It consists of one simple fully connected layer which takes 256 dimensionality combined features as input and produces one scores. $\theta_{m}=\{W_m, \mathbf{b}_{m}\}$ are the learnt parameters, where $W_m$ and $\mathbf{b}_{m}$ are the weight matrix and bias of the fully-connected layer, respectively.

\subsubsection{Objectness scorer}
Each node of the tree is related to the semantic information of the corresponding region, i.e., the semantic features. Objectness scorer $F_{o}$ directly predicts objectness scores in semantic feature space.
\begin{equation}
\begin{aligned}
o_{i}=F_{o}\left(\mathbf{x}_{i};\theta_{o}\right)=\phi\left(W_{o\_1}\sigma\left(W_{o\_0}\mathbf{x}_{i}+\mathbf{b}_{o\_0}\right)+\mathbf{b}_{o\_1}\right),
\label{eqn:objectness_scoring}
\end{aligned}
\end{equation}
where $\phi\left(\cdot\right)$ is the softmax operation. Our approach rejects candidate proposals that have low scores without compromising the recall rate. We experimentally found that one fully connected layer (merely consisting of 512 parameters) is so simple that it can not well fit thousands of proposals. Thus, we utilize two stacked fully connected layers to implement the objectness scorer, in which the first one is 256 to 256, followed by the rectified linear unit, and the second one is 256 to 2, followed by a softmax layer for objectness prediction. $\theta_{o}=\{W_{o\_0},W_{o\_1}, \mathbf{b}_{o\_0},\mathbf{b}_{o\_1}\}$ are the learnt parameters, where $W_{o\_0}$ and $\mathbf{b}_{o\_0}$ are the weight matrix and bias of the first fully-connected layer, while $W_{o\_1}$ and $\mathbf{b}_{o\_1}$ are those of the second one.


\subsection{Randomized Merging Algorithm}
\begin{figure}[htp]
\centering
\includegraphics[width=1.0\linewidth]{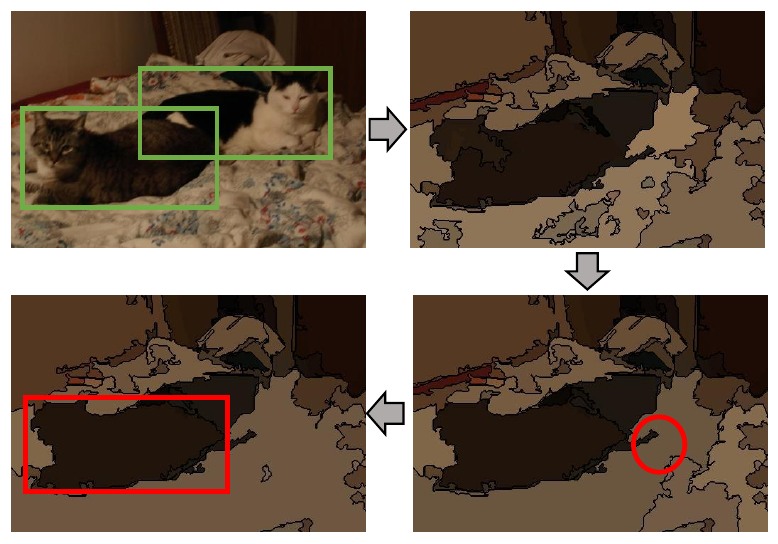}
\caption{An example of incorrect merging using the greedy merging algorithm. Top left: Input image; top right: over-segmentation; bottom right: incorrect merging; bottom left: merging result. The black-white cat is lost because its white part incorrectly merges with the background.}
\label{fig:greedy_merging}
\end{figure}

As discussed above, greedy merging groups the neighbouring regions with the highest similarity score for each iteration. Once a segment of an object mistakenly merges with a neighboring segment that belongs to surrounding objects or background, this object would have little chance to be found. Figure \ref{fig:greedy_merging} presents an example as an illustration. Given an image with a brown cat and a black-white one, the brown cat is successfully detected using the greedy merging processing, as it is distinguishable from the background (red bounding box in Figure \ref{fig:greedy_merging}). However, the white segment of the other cat incorrectly merges with a piece of background as they have more similar appearance (red circle in Figure \ref{fig:greedy_merging}). In this case, the subsequent merging process misses this cat inevitably. We propose a randomized merging algorithm to alleviate this problem. Instead of merging the neighbouring regions with the highest similarity score for each iteration, our approach selects one pair to merge among the top $k$ highest pairs according to a distribution constructed based on their scores. The randomized merging process can be repeated for several times to increase the diversity of the generated proposals. This helps to recall more incorrectly merged objects, as explained in Section \ref{subsec:Randomized_Inference}.

The randomized merging algorithm works as follows. Starting from the semantic features $\{\mathbf{x}_i\}_{i=1}^{N_{seg}}$ and over-segmented regions $\mathcal{R}=\{r_i\}_{i=1}^{N_{seg}}$ where $N_{seg}$ is the number of segments, our approach first computes the merging scores of all neighbouring regions using the feature combiner and merging scorer. Our approach then re-ranks the merging scores to obtain the $k$ pairs of neighbouring regions $\{\left(r_{i_t},r_{j_t}\right)\}_{t=1}^{k}$ with the top-$k$ highest scores $\{s_{i_t,j_t}\}_{t=1}^{k}$, and further constructs a multinomial probability distribution according to the $k$ merging scores,  expressed as
\begin{equation}
\begin{aligned}
\left(i_t,j_t\right)\sim{Mult}\left(\rho\right),
\label{eqn:mult}
\end{aligned}
\end{equation}
where
\begin{equation}
\begin{aligned}
\rho_{i_t,j_t}=\frac{\exp\left({s_{i_t,j_t}}\right)}{\sum_{t=1}^{k}\exp\left({s_{i_t,j_t}}\right)}, t=1,2,\cdots,k,
\label{eqn:probability}
\end{aligned}
\end{equation}
where $\rho_{i_t,j_t}$ indicates the probability that the $t$-th pair of regions can be selected.
Our approach randomly draws one pair of regions $\left(r_{i_{t'}},r_{j_{t'}}\right)$ according to the probability distribution ${Mult}\left(\rho\right)$, merges these two regions together, and then computes new merging scores between the resulting region and its neighbours. The process is repeated until the whole image becomes one region. The general process is detailed in Algorithm \ref{alg:random_greedy}. As the candidate object proposals, we consider the bounding boxes that tightly enclose the segments throughout the hierarchy. Then the objectness scores, learned by the objectness scorer, are used to rank the candidate proposals and the ones with low scores are rejected to get a certain number of proposals.

\begin{algorithm}[!t]
\caption{Randomized merging algorithm}
\begin{algorithmic}[1]
\REQUIRE
Initial region set $\mathcal{R}=\{r_i\}_{i=1}^{N_{seg}}$
\ENSURE
Set of object proposal $\mathcal{P}$

\STATE Initialize merging score set $\mathcal{S}=\emptyset$
\FOR {all neighbouring region pair $\left(r_i,r_j\right)$}          
\STATE Calculate merging score $s_{i,j}$
\STATE $\mathcal{S}=\mathcal{S}\cup{s_{i,j}}$
\ENDFOR
\WHILE{$\mathcal{S}\neq\emptyset$}
\STATE Get the $k$ highest merging scores $\{s_{i_t,j_t}\}_{t=1}^{k}$
\STATE Construct multinomial distribution ${Mult}\left(\rho\right)$
\STATE Select randomly $t'$-th pair according to ${Mult}\left(\rho\right)$
\STATE Merge corresponding regions $r_{t'}={r_{i_{t'}}}\cup{r_{j_{t'}}}$
\STATE Remove scores regarding ${r_{i_{t'}}}$: $\mathcal{S}=\mathcal{S} \setminus s_{{i_{t'}},\ast}$
\STATE Remove scores regarding ${r_{j_{t'}}}$: $\mathcal{S}=\mathcal{S} \setminus s_{{j_{t'}},\ast}$
\STATE Compute merging score set $\mathcal{S}_{t'}$ between $r_{t'}$ and its neighbours
\STATE Update merging score set $\mathcal{S}=\mathcal{S} \cup \mathcal{S}_{t'}$
\STATE Update region set $\mathcal{R}=\mathcal{R} \cup r_{t'}$
\ENDWHILE
\STATE Extract object proposals $\mathcal{P}$ from all regions in $\mathcal{R}$
\end{algorithmic}
\label{alg:random_greedy}
\end{algorithm}

\section{Optimization}
\label{sec:optimization}
Suppose that we have the training set $\mathcal{X}=\{\left(I_i,c_i,b_i\right)|i=1,2,...,N\}$, where $N$ is the number of training samples; $I_i$ is the $i$-th input sample, including the local features of all regions and the adjacency matrix (as shown in Figure \ref{fig:tree}(a) and (b)); $c_i$ and $b_i$ are the corresponding class labels of regions and ground truth object bounding boxes, respectively. Our model is jointly trained with two objectives: 1)~the merging loss $\mathcal{L}_{m}$ penalizes incorrect region grouping in the hierarchical tree structure; and 2)~the objectness loss $\mathcal{L}_{o}$ helps to learn the objectness scorer. Therefore, we define the structured loss as

\begin{equation}
\begin{aligned}
\mathcal{L}=\mathcal{L}_{m}+\lambda\mathcal{L}_{o}+\frac{\eta}{2}{||\theta||}^2_2,
\label{eqn:loss}
\end{aligned}
\end{equation}
where ${\theta}=\{{\theta}_{s},{\theta}_{c},{\theta}_{m},{\theta}_{o}\}$ are the set of parameters to learn and ${||\theta||}^2_2$ is the L2 norm regularization term. $\lambda$ and $\eta$ are two balance parameters.

\subsection{Merging Loss}
\begin{figure}[htp]
\centering
\includegraphics[width=0.95\linewidth]{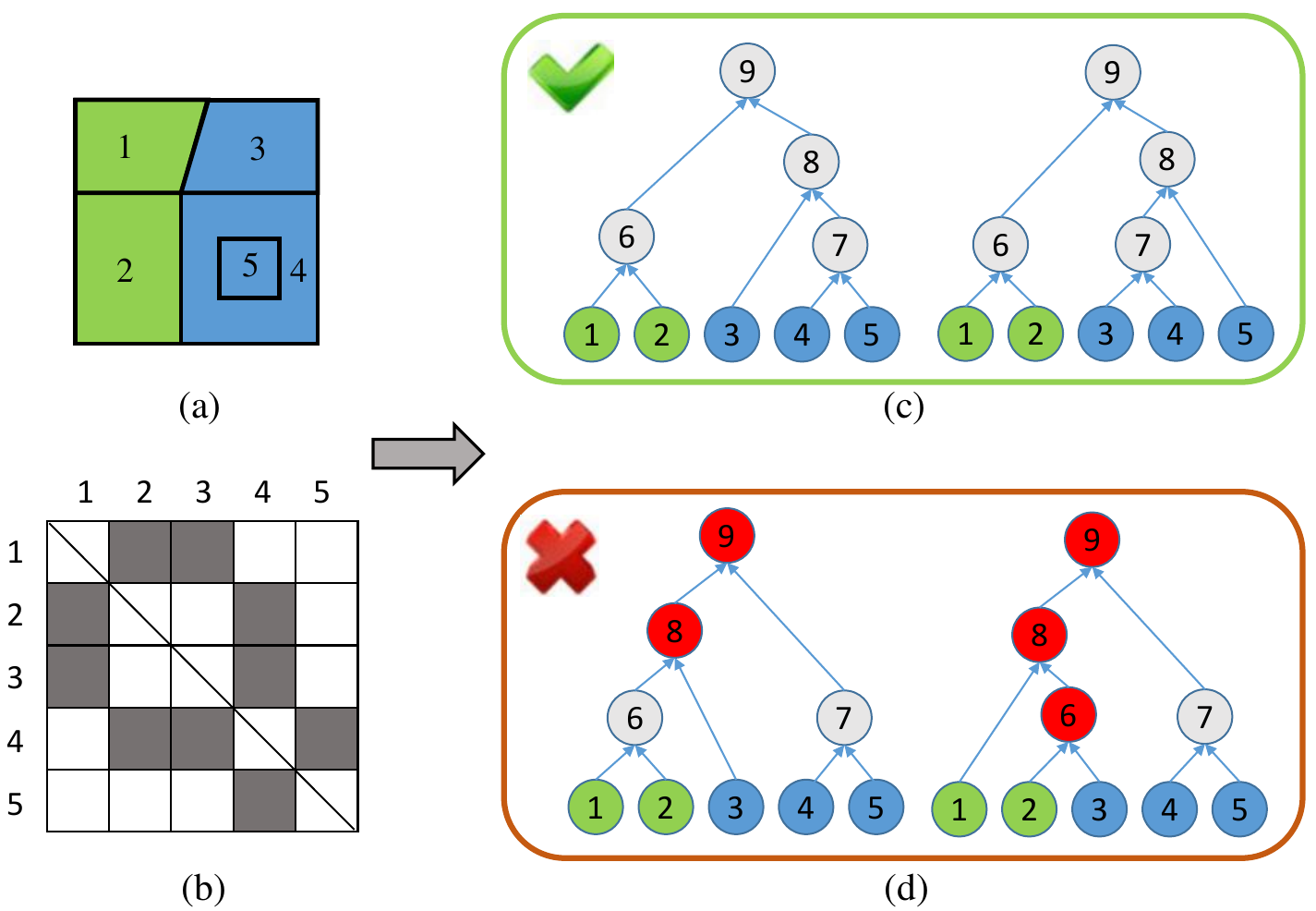}
\caption{Examples of generating correct and incorrect trees. (a) Input image, green and blue indicate differently labelled regions. (b) Adjacent matrix of image regions; (c) correct trees; (d) incorrect trees.}
\label{fig:tree}
\end{figure}

Given an input image $I$, its bottom-up merging process can be presented as $RN\left(\theta,I,t\right)$, and it produces a binary tree $t\in\mathcal{T}\left(I\right)$, where $\mathcal{T}\left(I\right)$ is the set of all possible binary trees constructed from input $I$. In the learning stage, the class labels of all the segmented regions are available. We further define $\mathcal{T}\left(I,c\right)$ as the set of all possible correct trees. Here, a tree is regarded as correct if any region merges with the one belonging to the same class before other regions from different classes. Figure \ref{fig:tree} presents some examples of generating correct and incorrect trees from an image.

Inspired by ~\cite{socher2011parsing,taskar2004max}, we define a margin loss function $\triangle{L}:{\mathcal{I}}\times{\mathcal{C}}\times{\mathcal{T}}\rightarrow{\mathbb{R}^{+}}$, where $\triangle{L\left(I,c,t\right)}$ measures the penalty of the construction of a parsing tree $t$ for input $I$ with label $c$. In the context of recursive merging process, the loss increases when a segment merges with the one from different class before those with the same class label. We denote $N\left(t\right)$ as the set of non-terminal nodes of tree $t$, and $subtree\left(d\right)$ as a subtree underneath the non-terminal node for each $d\in{N\left(t\right)}$. Naturally, we formulate the loss by penalizing the incorrect subtrees

\begin{equation}
\begin{aligned}
\triangle{L\left(I,c,t\right)}=\sum_{d\in{N\left(t\right)}}{\textbf{1}\{subtree\left(d\right)\notin{\mathcal{T}\left(I,c\right)}}\},
\label{eqn:margin_loss}
\end{aligned}
\end{equation}
where $\textbf{1}\left\{\cdot\right\}$ is an indicator function whose value is 1 when the expression is true and 0 otherwise. Figure \ref{fig:tree}(d) illustrates two examples of incorrect trees, in which the margin losses are 2 and 3, respectively.

Our goal is to learn a function $f_{\theta}(\cdot)$ with small expected loss on the unseen inputs. Similar to ~\cite{socher2011parsing,taskar2004max}, we consider the following forms
\begin{equation}
\begin{aligned}
{f_{\theta}\left(I\right)}=\argmax_{t\in{\mathcal{T}\left(I\right)}}\left\{s\left(RN\left(\theta,I,t\right)\right)\right\},
\label{eqn:expected_function}
\end{aligned}
\end{equation}
where $s\left(\cdot\right)$ predicts the score for a tree by summing up merging scores of all the merged neighbouring pairs. In the optimization procedure, we aim to learn a score function that assigns higher scores to correct trees than incorrect ones. Given the parameters $\theta$, we first define the margin between the correct tree $t_i$ and another tree $t$ for $I_i$,
\begin{equation}
\begin{aligned}
s\left(RN\left(\theta,I_i,t_i\right)\right)-s\left(RN\left(\theta,I_i,t\right)\right).
\label{eqn:margin_dif}
\end{aligned}
\end{equation}
Intuitively, the margin will be enlarged as the margin loss function $\triangle{L\left(I,c,t\right)}$ increases, expressed as
\begin{equation}
\begin{aligned}
s\left(RN\left(\theta,I_i,t_i\right)\right)-s\left(RN\left(\theta,I_i,t\right)\right)\geq{\kappa\triangle{L\left(I,c,t\right)}},
\label{eqn:margin_inequation}
\end{aligned}
\end{equation}
where $\kappa$ is a parameter. The merging loss can be thus defined as

\begin{equation}
\begin{aligned}
\mathcal{L}_{m}=\sum_{i=1}^N{\mathcal{L}_{m}^{(i)}},
\label{eqn:total_structure_loss}
\end{aligned}
\end{equation}
where

\begin{equation}
\begin{aligned}
{\mathcal{L}_{m}^{(i)}}=&\max_{t\in{\mathcal{T}\left(I_i\right)}}\left\{s\left(RN\left(\theta,I_i,t\right)\right)+{\kappa\triangle{L\left(I_i,c_i,t\right)}}\right\} \\
&-\max_{t_i\in{\mathcal{T}\left(I_i,c_i\right)}}\left\{s\left(RN\left(\theta,I_i,t_i\right)\right)\right\}.
\label{eqn:structure_loss}
\end{aligned}
\end{equation}
Optimizing the merging loss can maximize the correct trees' scores while minimizing the scores of the highest scoring but incorrect trees. Following \cite{socher2011parsing}, we utilize the greedy merging to approximatively find an tree with maximum scores among $\mathcal{T}(I_i)$, and a correct tree with maximum scores among $\mathcal{T}(I_i,c_i)$. The gradients are computed and back propagated based on these two selected trees.

\subsection{Objectness Loss}
One of the main advantages of our approach is that it can simultaneously predicts an objectness score for each proposal candidate, which can be used for proposal ranking and rejecting the ones with low scores. We simply employ a softmax classifier with the semantic features of each node. We generate positive and negative samples from all of the regions as follows. Given a region, we first calculate the IoU scores between the box that tightly bounds this region with each ground truth bounding box. If the maximum IoU is larger than 0.5, this region is considered as positive; and if the maximum IoU is smaller than 0.2, it is used as a negative sample. All these regions are considered as useful regions to define the objectness loss. We simply ignore other regions since they may not provide discriminative information. For the $i$-th useful region, the loss function can be defined as

\begin{equation}
\begin{aligned}
\mathcal{L}_{o}^{(i)}=-\sum_{l=0}^{1}{\textbf{1}\left\{l_{i}=l\right\}\log\left(p_{i,l}\right)},
\label{eqn:softmax_loss}
\end{aligned}
\end{equation}
where $p_{i,l}$ is the score corresponding to the likelihood of the region belonging to label $l$. Hence

\begin{equation}
\begin{aligned}
\mathcal{L}_{o}=\sum_{i=1}^{N_u}{\mathcal{L}_{o}^{(i)}},
\label{eqn:total_softmax_loss}
\end{aligned}
\end{equation}
where $N_u$ is the number of useful regions.

The model is jointly trained by the stochastic gradient descent (SGD) with momentum ~\cite{bottou2012stochastic}.

\section{Experiment}
In this section, we present the extensive experimental results to compare with state-of-the-art methods, demonstrating the superiority of the proposed methods, and analyze the benefit of introducing the randomized merging algorithm for object proposals generation.

\label{sec:experiment}
\subsection{Experimental Setting}

\subsubsection{Datasets}
We first conduct the experiments on the PASCAL VOC2007 dataset ~\cite{everingham2010pascal}, which consists of 9,963 images from 20 categories of objects. The model is trained using 422 images of the PASCAL VOC2007's segmentation set. We compare the performance of our approach with those of state-of-the-art methods, and evaluate the contribution of randomized merging algorithm using the 4,952 test images that contain 14,976 objects, including the ``difficult" ones. To better demonstrate the effectiveness of the proposed method, we also conducted experiments on the PASCAL VOC 2012 validation set, which contains 15,787 objects in 5,823 images. As our model is trained with 20 object categories on PASCAL VOC, we further investigate the generalization ability of our method to unseen object categories on ImageNet 2015 validation dataset ~\cite{russakovsky2014imagenet}, which contains about 20,000 images of 200 categories, without re-training the model using the training samples from ImageNet.

\subsubsection{Evaluation Metrics}
One of the primary metrics is the Intersection over Union (IoU) measure, where the IoU is defined as the intersection area of the proposal, and the ground truth bounding box divided by their union area. For a fixed number of proposals, the recall rate (the fraction of ground truth annotations covered by proposals) varies as the IoU threshold increases from 0.5 to 1, so that a recall-IoU curve can be obtained. Besides, the curves indicating the recall rate with reference to the number of ranked proposals, are also given, with IoU fixed as both 0.5 and 0.8, respectively. This is widely adopted by many proposal works ~\cite{hosang2015makes, wang2015object} for evaluation. We also compare the average recall (AR), defined as the average recall when IoU ranges from 0.5 to 1 ~\cite{hosang2015makes,chen2015improving}, since AR is considered to be strongly correlated with detection performance.

\begin{figure*}[!t]
\centering
{\includegraphics[width=0.88\linewidth]{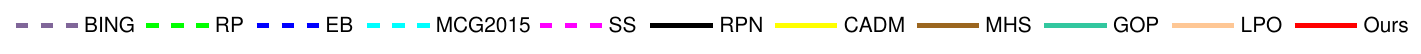}}
\vspace{6pt}
\subfigure[]{
\label{fig:subfig1_1} 
\includegraphics[width=0.315\linewidth]{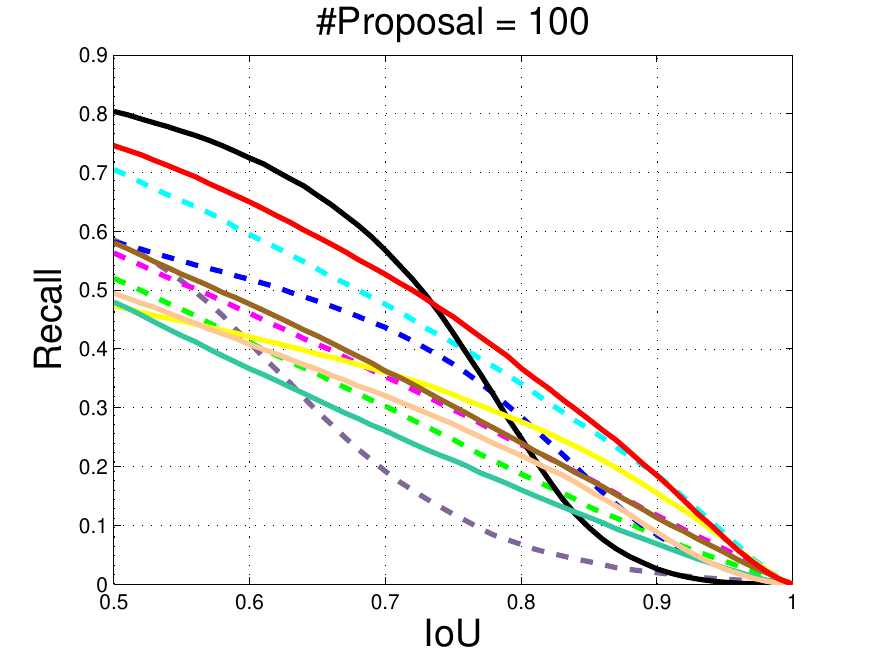}}
\subfigure[]{
\label{fig:subfig1_2} 
\includegraphics[width=0.315\linewidth]{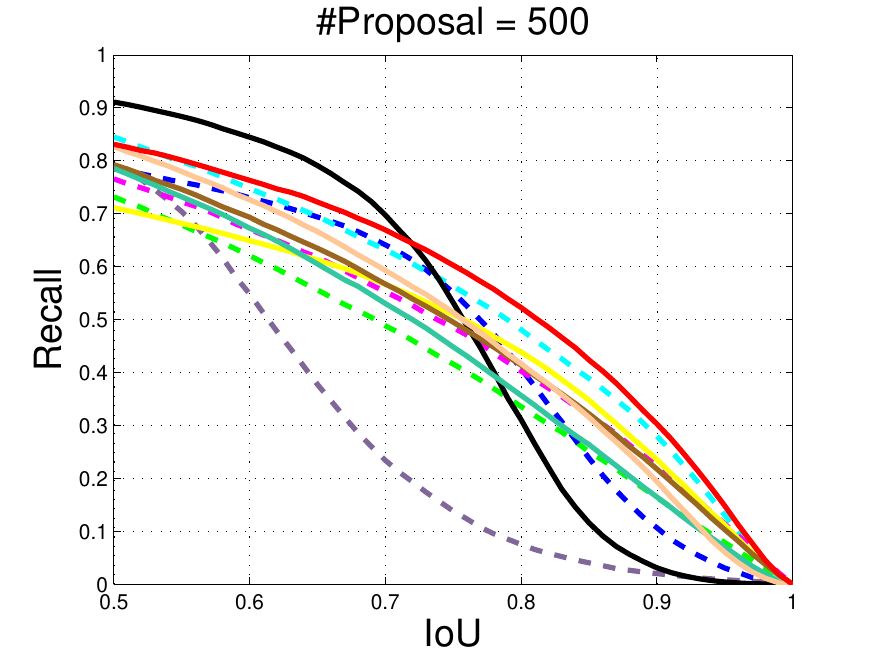}}
\subfigure[]{
\label{fig:subfig1_3} 
\includegraphics[width=0.315\linewidth]{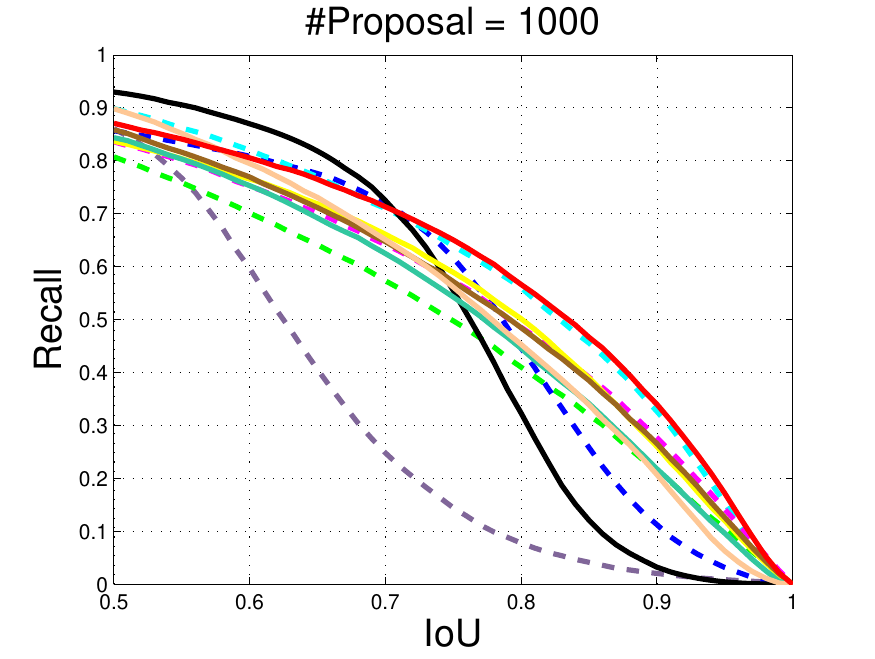}}
\subfigure[]{
\label{fig:subfig1_4} 
\includegraphics[width=0.315\linewidth]{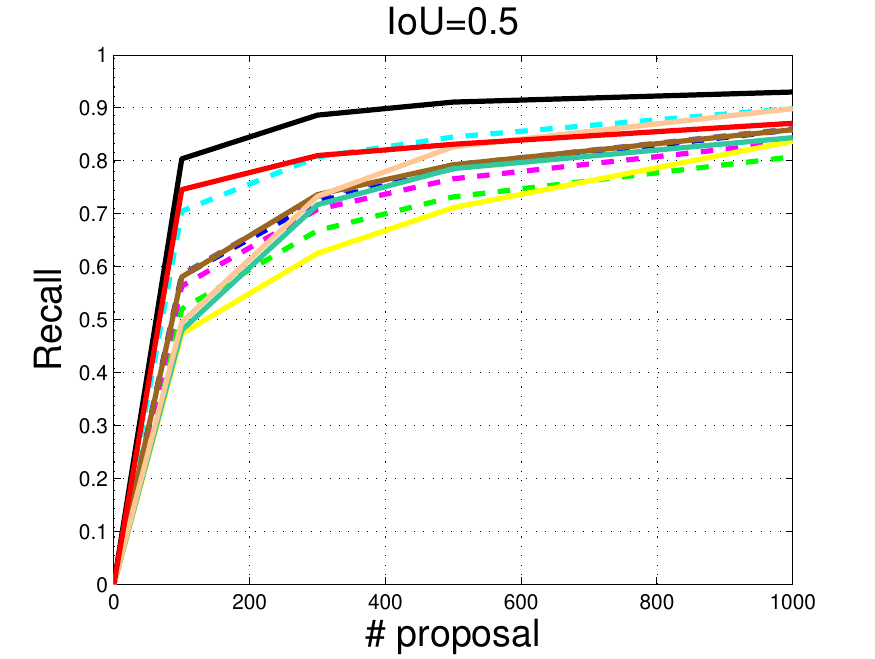}}
\subfigure[]{
\label{fig:subfig1_5} 
\includegraphics[width=0.315\linewidth]{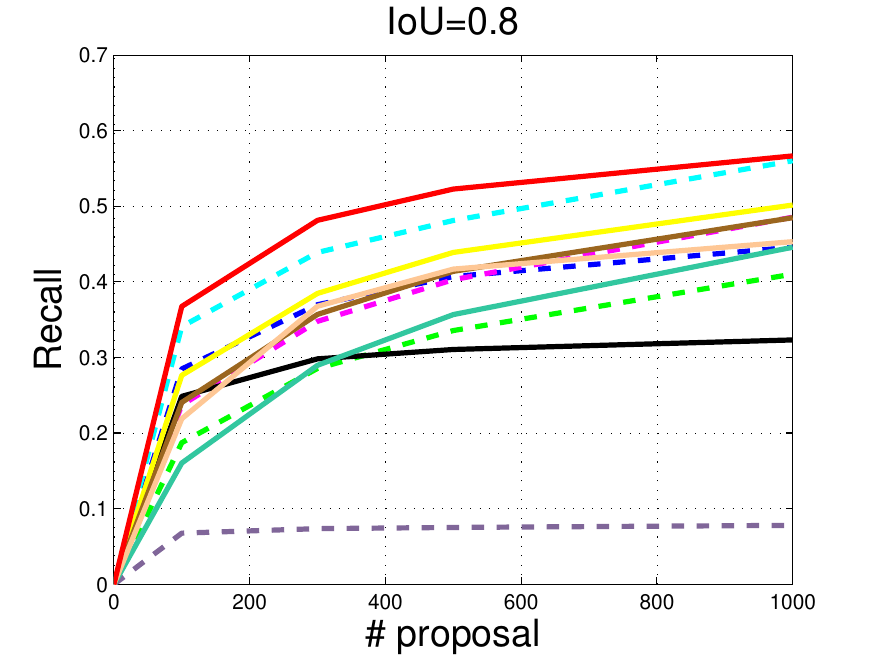}}
\subfigure[]{
\label{fig:subfig1_6} 
\includegraphics[width=0.315\linewidth]{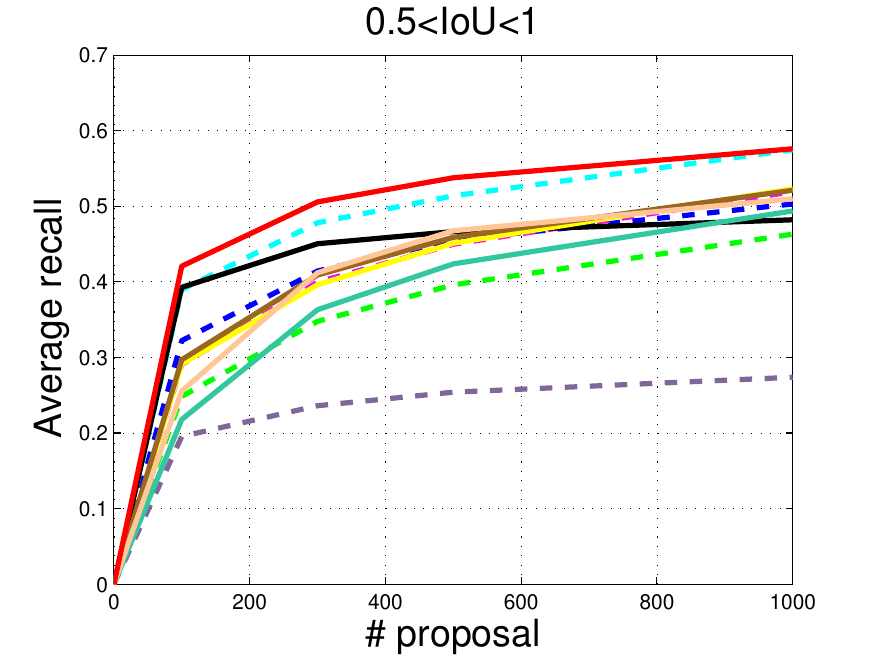}}
\caption{Comparison of our proposed method and other state-of-the-art approaches on the PASCAL VOC 2007 test set. Best viewed in color.}
\label{fig:VOC2007_result1}
\end{figure*}

\begin{figure*}[!t]
\centering
{\includegraphics[width=0.72\linewidth]{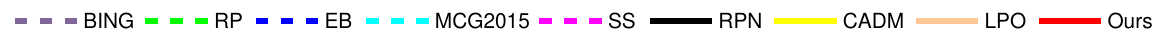}}
\vspace{6pt}
\subfigure[]{
\label{fig:subfig2_1} 
\includegraphics[width=0.315\linewidth]{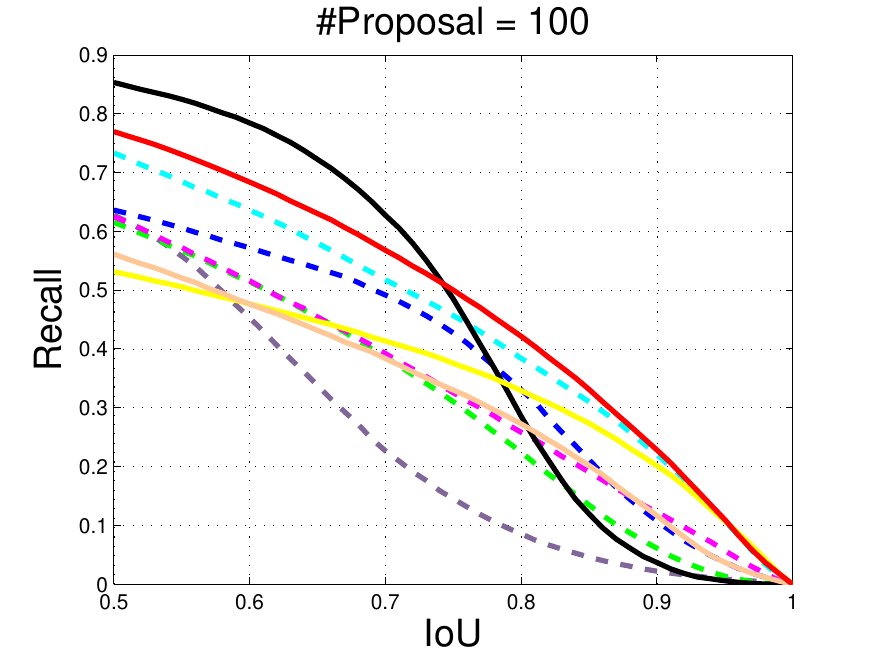}}
\subfigure[]{
\label{fig:subfig2_2} 
\includegraphics[width=0.315\linewidth]{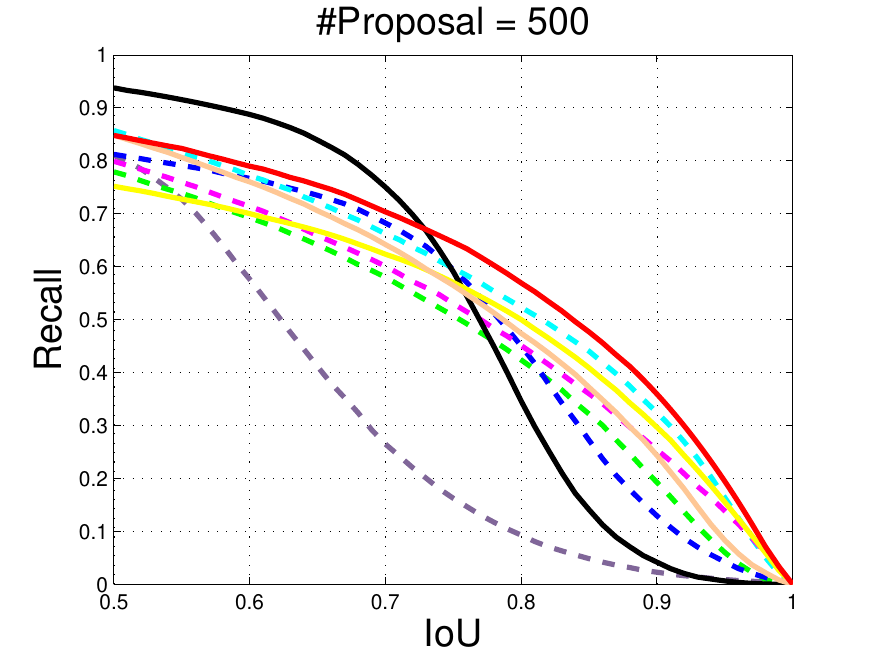}}
\subfigure[]{
\label{fig:subfig2_3} 
\includegraphics[width=0.315\linewidth]{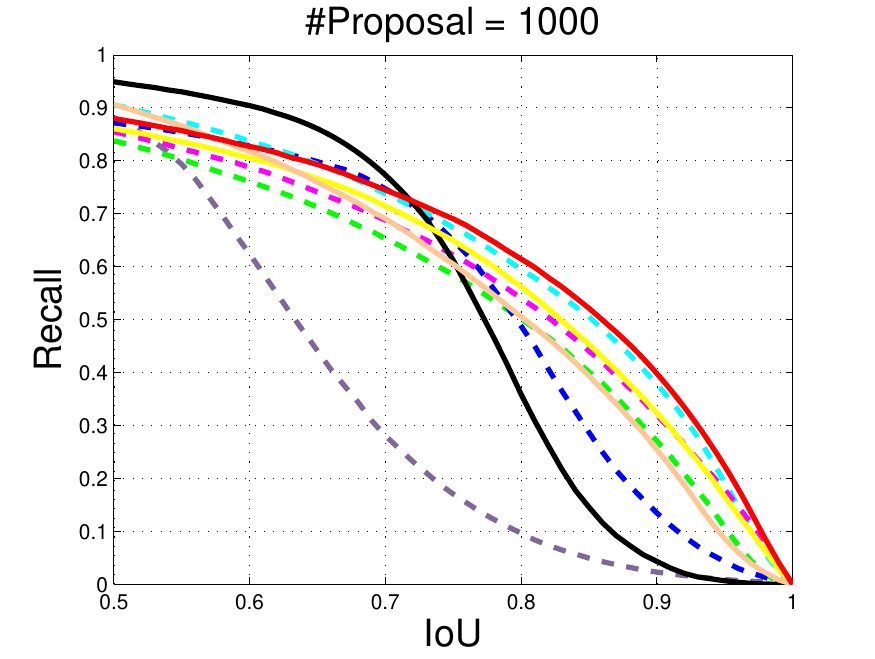}}
\subfigure[]{
\label{fig:subfig2_4} 
\includegraphics[width=0.315\linewidth]{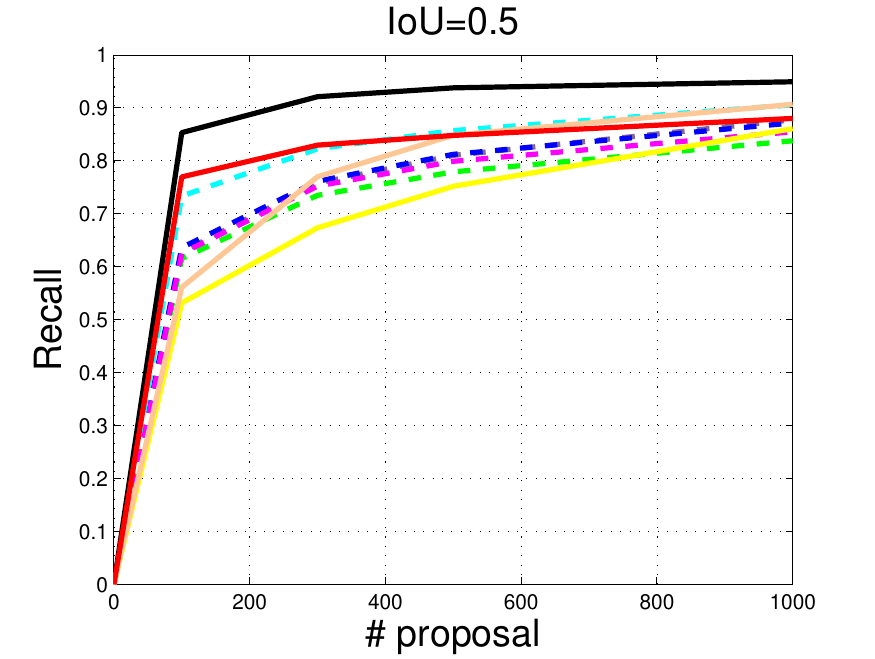}}
\subfigure[]{
\label{fig:subfig2_5} 
\includegraphics[width=0.315\linewidth]{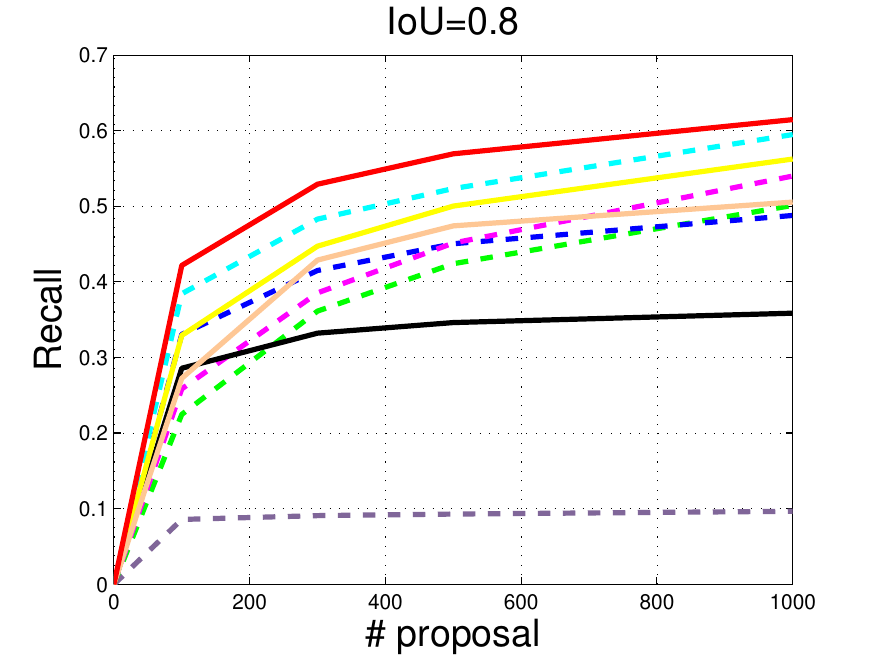}}
\subfigure[]{
\label{fig:subfig2_6} 
\includegraphics[width=0.315\linewidth]{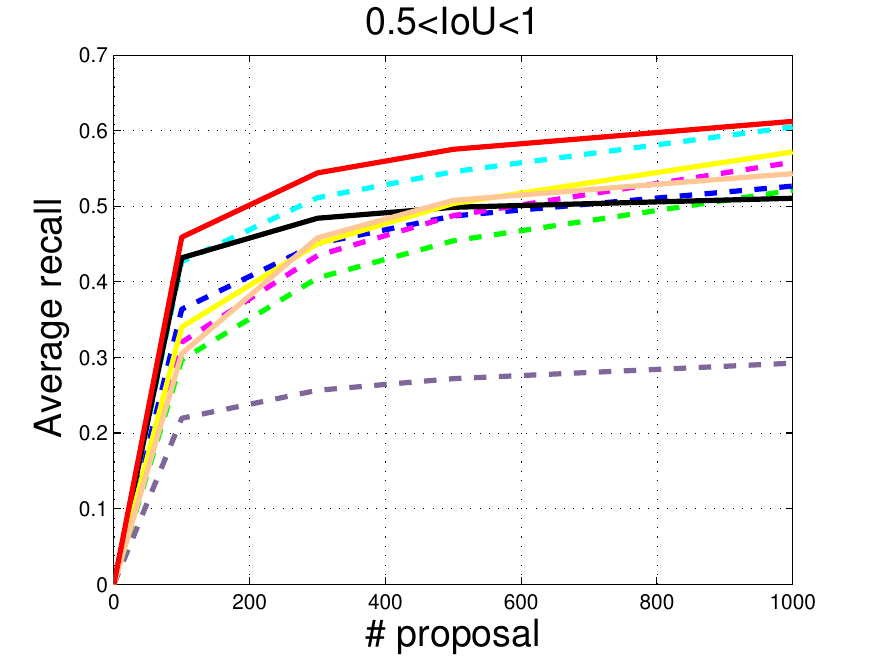}}
\caption{Comparison of our proposed method and other state-of-the-art approaches on the PASCAL VOC 2012 validation set. Note that our model is trained on PASCAL VOC 2007, but still achieves the best performance against other competitors. Best viewed in color.}
\label{fig:VOC2012_result1}
\end{figure*}

\subsubsection{Implementation Details}
Following ~\cite{uijlings2013selective}, we adopte the efficient graph-based method ~\cite{felzenszwalb2004efficient} to produce initial over-segmentations with four parameter values (i.e., $k=100,150,200,250$), respectively. We implement the proposed model using Caffe open source library ~\cite{jia2014caffe}, and train it by stochastic gradient descent (SGD) with a batch size of 2, momentum of 0.9, weight decay of 0.0005. The learning rate is initialized as $10^{-5}$ and divided by 10 after 20 epochs. The balance parameter $\lambda$ in Equation \ref{eqn:loss} is simply set as 1.  Note that it is indeed possible to perform joint training for the fast RCNN and the ReNN. We do not train the model in this way, because only 422 images are provided for training, which easily leads to over-fitting. Therefore, we simply use the fast RCNN model pre-trained on the PASCAL VOC 2007 detection dataset for local feature extraction, and then train our ReNN model alone. In the random merging algorithm, the parameter $k$ is set as 5. To improve the quality of generated proposals, we perform the random merging process for $K$ times ($K=8$ in our experiments), and rank all the generated proposals using the objectness scores. The proposals with low scores are rejects to get a certain number of proposals for evaluation.

\begin{figure*}[!t]
\centering
\includegraphics[width=1.0\linewidth]{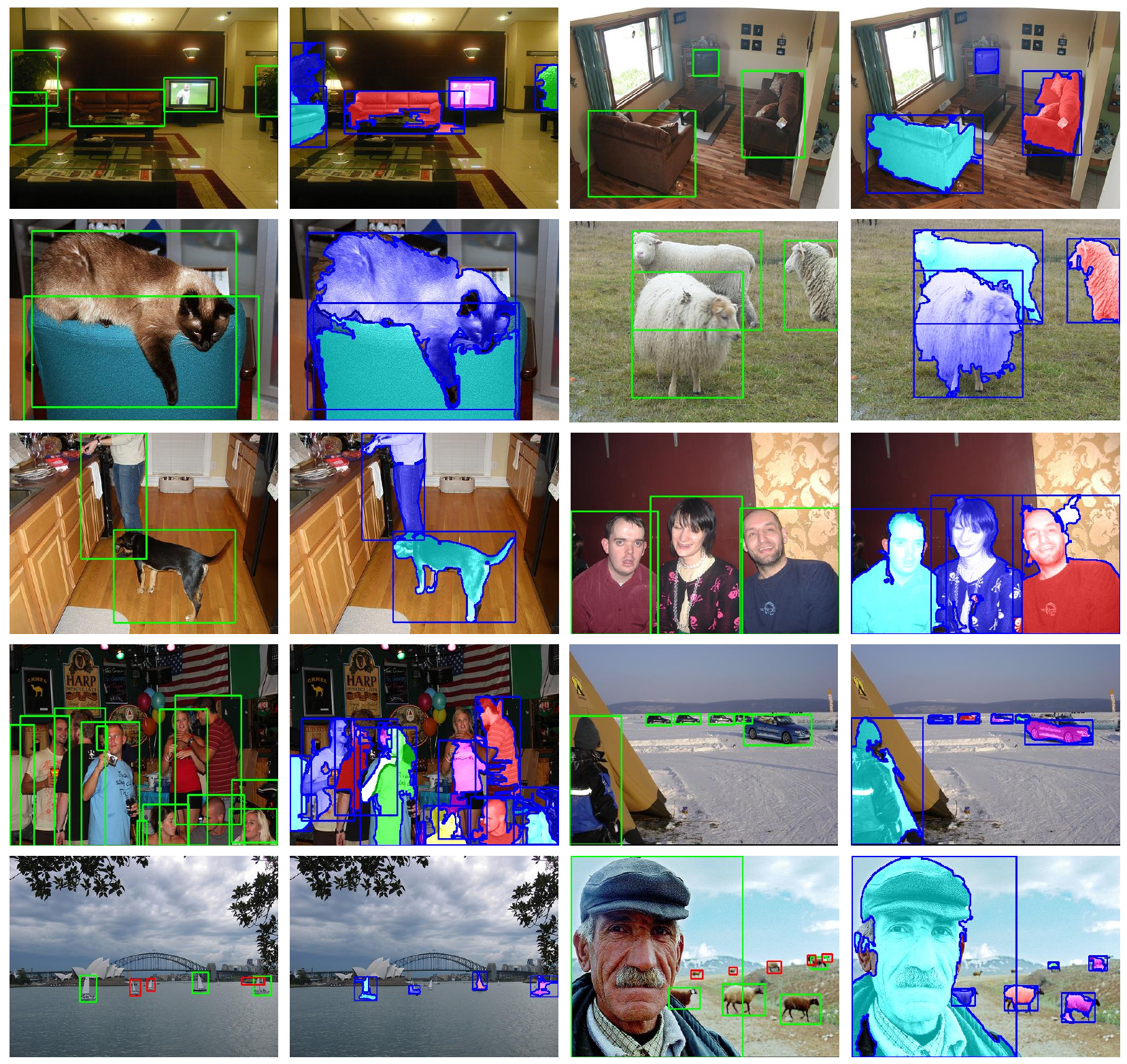}
\caption{Qualitative examples of our object proposals. Ground truth boxes are shown in green and red, with green indicating the object is found and red indicating the object is not found. The blue boxes are the object proposals with highest IoU to each ground truth box, and the blue silhouettes are the corresponding object contours. All the samples are taken from PASCAL VOC dataset.}
\label{fig:visualization_results2}
\end{figure*}

\subsection{Comparison with State-of-the-Art Approaches}
In this subsection, we compare our method with recent state-of-the-art methods, including BING ~\cite{cheng2014bing}, Randomized Prim (RP) ~\cite{manen2013prime}, EdgeBox (EB) ~\cite{zitnick2014edge}, Multiscale Combinatorial Grouping (MCG2015) \footnote{MCG2015 is the improved version of original MCG and achieves much better performance.} ~\cite{pont2015multiscale}, Selective Search (SS) ~\cite{uijlings2013selective}, Faster R-CNN (RPN) ~\cite{ren2015faster}, Complexity Adaptive Distance Metric (CADM) ~\cite{xiao2015complexity}, Multi-branch Hierarchical Segmentation (MHS)\footnote{We only compare with MHS on the PASCAL VOC 2007 dataset, because only the results on this dataset are available.}~\cite{wang2015object}, Geodesic Object Proposals (GOP) \footnote{We only compare with GOP on the PASCAL VOC 2007 and ImageNet datasets, because only the results on these two datasets are available.}~\cite{krahenbuhl2014geodesic}, Learn to Propose Objects (LPO) ~\cite{krahenbuhl2015learning}. 
In our experiments, we use Edgebox70 (optimal settings for an IoU threshold of 0.7) for EB, and default settings for others, in order to ensure the best overall performance for these methods. In addition, we follow ~\cite{hosang2015makes} to control the number of candidates to a specific value for a fair comparison. Since BING, MCG2015, SS, CADM, RPN, EB, MHS and GOP provide sorted proposals, we select $n$ proposals with top $n$ highest scores for evaluation. However, RP and LPO does not provide the scores to rank the proposals, so we simply select the first $n$ proposals in our experiments.

We first analyze the experimental results on the PASCAL VOC 2007 dataset, as depicted in Figure \ref{fig:VOC2007_result1}. It can be observed that window-scoring-based methods (e.g., EB and RPN) achieve competitive recall rates with a relatively low IoU threshold. This mainly benefits from the exhaustive search over locations and scales, and the accuracy of rejecting the non-objects by the window-scoring-based methods. However, their recall rates drop significantly when the IoU threshold increases. In contrast, region-grouping-based methods yield better performance as the IoU threshold increases. It is shown that MCG2015 performs best among region-grouping-based methods, but it is very time-consuming (over 30s per image) and may not practical especially for real-time object detection systems. It is noteworthy that our method runs $7\times$ faster than MCG2015, and meanwhile outperforms MCG2015 overall, particularly when the number of object proposals is strictly constrained (e.g., with 100 or 500 proposals).

\begin{table*}[htp]
\centering
\begin{tabular} 
{p{1.2cm}|p{0.3cm}p{0.3cm}p{0.3cm}p{0.3cm}p{0.4cm}p{0.3cm}p{0.20cm}p{0.20cm}p{0.3cm}p{0.3cm}p{0.3cm}p{0.3cm}p{0.35cm}p{0.4cm}p{0.45cm}p{0.3cm}p{0.35cm}p{0.3cm}p{0.3cm}p{0.5cm}|p{0.5cm}}
\hline
\centering Methods  & aero & bike & bird & boat & bottle & bus & car & cat & chair & cow & table & dog & horse & mbike & person & plant & sheep & sofa & train & tv & mAP \\
\hline\hline
\centering SS      & \textbf{64.8} & 67.9 & 52.9 & 45.3 & 20.6 & 69.9 & 65.6 & 70.7 & 30.0 & 62.7 & 59.8 & 62.3 & 73.6 & 64.9 & 54.8 & 24.8 & 49.3 & 60.2 & 71.7 & 55.2 & 56.4 \\
\centering EB  & 62.7 & 68.0 & 52.5 & \textbf{45.8} & 24.6 & 65.2 & 69.2 & \textbf{70.8} & 29.8 & \textbf{64.2} & 59.3 & 61.7 & \textbf{74.9} & 67.5 & 59.5 & 26.8 & 51.6 & 55.2 & 71.1 & 56.9 & 56.9 \\
\centering MCG2015 & 64.2 & 70.6 & 50.3 & 42.5 & 26.4 & 70.8 & 66.4 & 69.3 & 29.8 & 63.7 & 61.2 & 61.0 & 72.7 & 65.5 & 57.7 & 28.1 & 50.3 & 61.1 & 70.3 & 59.1 & 57.0 \\
\centering RPN   & 61.5 & \textbf{71.2} & 53.5 & 42.6 & \textbf{29.0} & \textbf{72.3} & \textbf{72.5} & 70.7 & 32.2 & 63.5 & 56.0 & 62.8 & \textbf{74.9} & \textbf{68.2} & \textbf{62.4} & 27.6 & \textbf{53.9} & 53.6 & 66.7 & 58.5 & 57.7 \\
\hline
\centering Ours   & 64.6 & 67.1 & \textbf{56.1} & 44.0 & 27.7 & 70.1 & 66.7 & 70.5 & \textbf{35.9} & 60.5 & \textbf{61.8} & \textbf{64.0} & 74.2 & 66.1 & 57.7 & \textbf{30.2} & 52.6 & \textbf{62.7} & \textbf{73.2} & \textbf{65.7}   & \textbf{58.6} \\
\hline\hline
\centering SS & \textbf{74.5} & 77.7 & 66.6 & \textbf{58.3} & 32.8 & 77.9 & 76.0 & \textbf{83.8} & 43.3 & 75.9 & \textbf{70.2} & 77.7 & 80.0 & 75.7 & 67.0 & 32.3 & 64.3 & 66.6 & \textbf{78.1} & 65.3 & 67.2 \\
\centering EB  & 73.0 & \textbf{78.1} & 67.3 & 56.2 & 43.5 & \textbf{80.2} & 77.0 & 83.0 & 46.8 & 74.5 & 63.6 & 78.1 & 80.0 & \textbf{76.1} & 69.5 & 37.0 & 67.0 & 66.4 & 74.4 & 63.5 & 67.7\\
\centering MCG2015 &  74.5 & 77.9 & 66.6 & 52.5 & 42.2 & 78.4 & 77.9 & 80.0 & 45.1 & 71.7 & 66.4 & 77.5 & 80.1 & 74.1 & 72.4 & 34.0 & 67.5 & \textbf{68.0} & 77.6 & 67.7 & 67.6  \\
\centering RPN   & 72.9 & 77.7 & 69.2 & 57.3 & \textbf{44.8} & 78.4 & \textbf{81.7} & 82.5 & 44.5 & \textbf{77.3} & 62.5 & 80.4 & 81.4 & 74.6  & 73.6 & 33.6 & \textbf{71.3} & 65.5 & 77.1 & 64.4 & 68.5 \\
\hline
\centering Ours   & 72.6 & 77.3 & \textbf{71.6} & 54.0 & 40.1 & 78.5 & 78.2 & 83.6 & \textbf{47.6} & 74.9 & 67.3 & \textbf{81.3} & \textbf{83.6} & \textbf{76.1} & \textbf{74.5} & \textbf{39.3} & 64.6 & 65.5 & 76.7 & \textbf{72.4} & \textbf{69.0} \\
\hline
\end{tabular}
\vspace{2pt}
\caption{Comparison of object detection performance with proposals generated by different methods. All of the detectors are learned by the fast RCNN on PASCAL VOC 2007 trainval set, and tested with 20 categories on the PASCAL VOC 2007 test set. The upper part present the results using Fast R-CNN with CaffeNet and the lower part shows those using the Fast R-CNN with VGG-Net.}
\label{table:det}
\end{table*}

\begin{figure}[htp]
\centering
\includegraphics[width=0.8\linewidth]{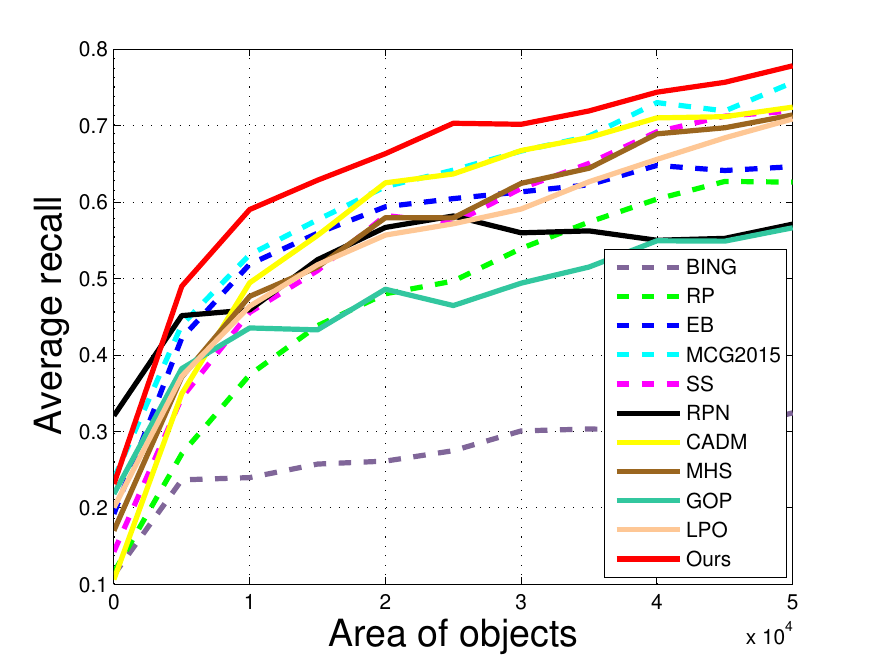}
\caption{Comparison of average recall (AR) with respect to different sizes of ground-truth objects on the PASCAL VOC 2007 test set. All of the AR rates are computed with top ranked 500 proposals per image. Best viewed in color.}
\label{fig:ar_area}
\end{figure}

\begin{figure*}[htp]
\centering
{\includegraphics[width=0.80\linewidth]{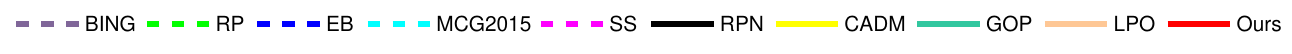}}
\vspace{6pt}
\subfigure[]{
\label{fig:subfigi2_1} 
\includegraphics[width=0.32\linewidth]{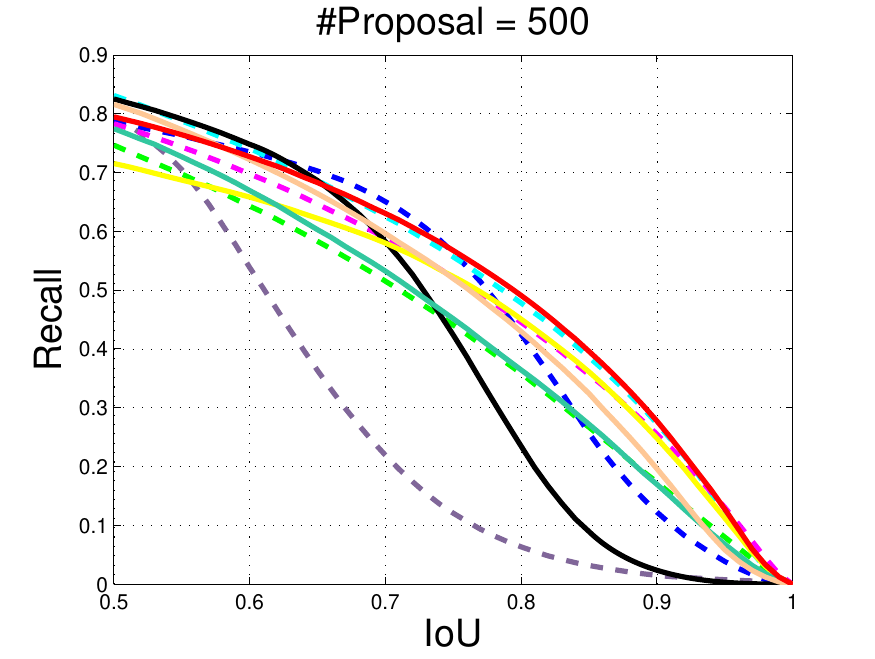}}
\subfigure[]{
\label{fig:subfigi2_2} 
\includegraphics[width=0.32\linewidth]{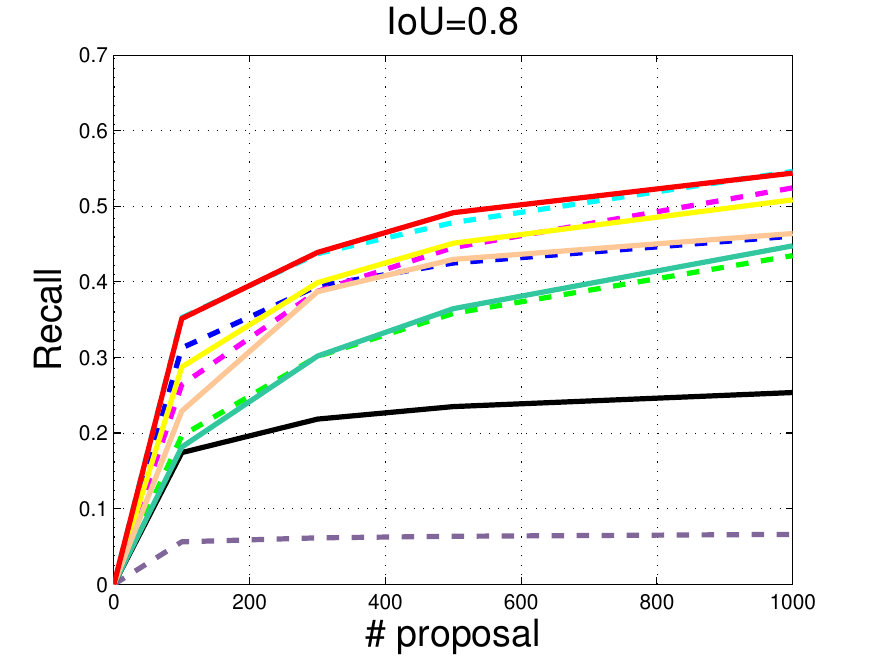}}
\subfigure[]{
\label{fig:subfigi2_3} 
\includegraphics[width=0.32\linewidth]{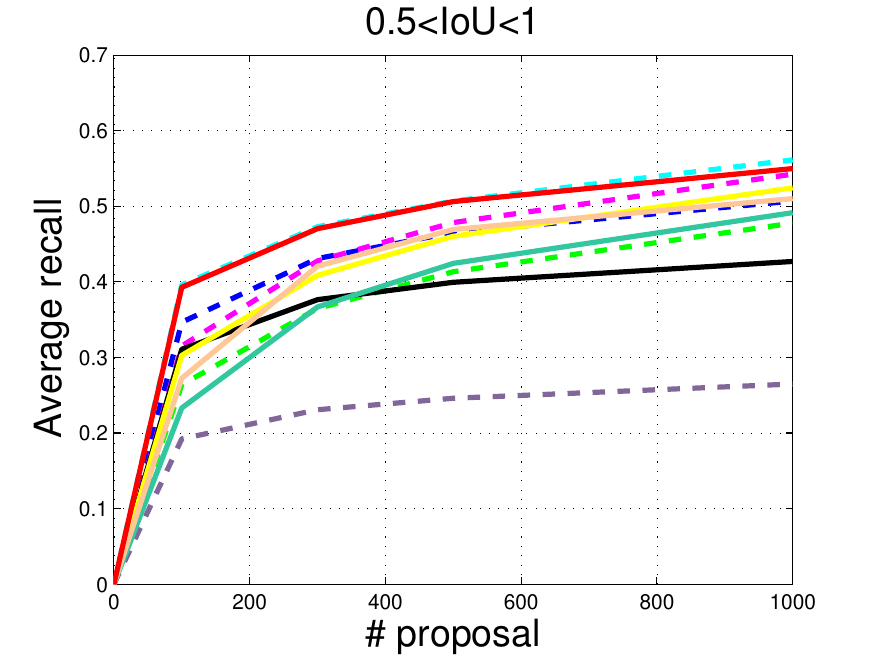}}
\caption{Comparison of our proposed method and other state-of-the-art approaches on the ImageNet 2015 validation set. Best viewed in color.}
\label{fig:ImageNet_result1}
\end{figure*}

\begin{figure*}[!t]
\centering
\subfigure[]{
\label{fig:randomness_ar} 
\includegraphics[width=0.32\linewidth]{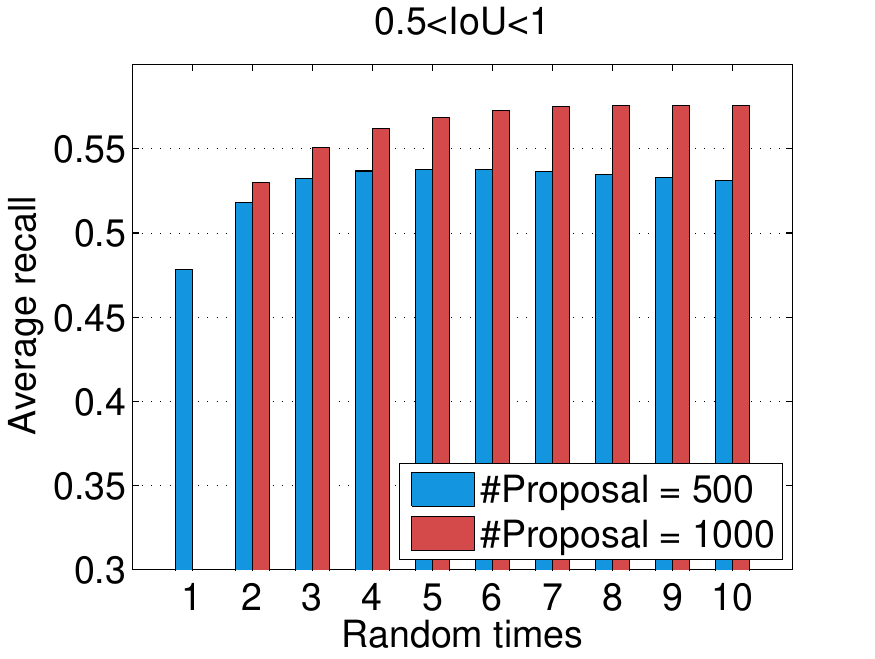}}
\subfigure[]{
\label{fig:randomness_05R} 
\includegraphics[width=0.32\linewidth]{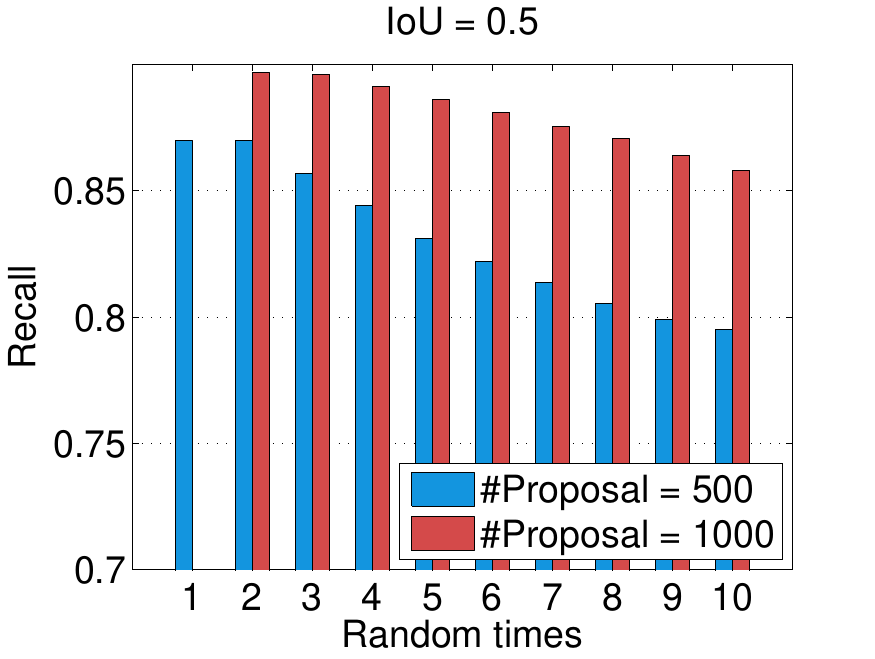}}
\subfigure[]{
\label{fig:randomness_08R} 
\includegraphics[width=0.32\linewidth]{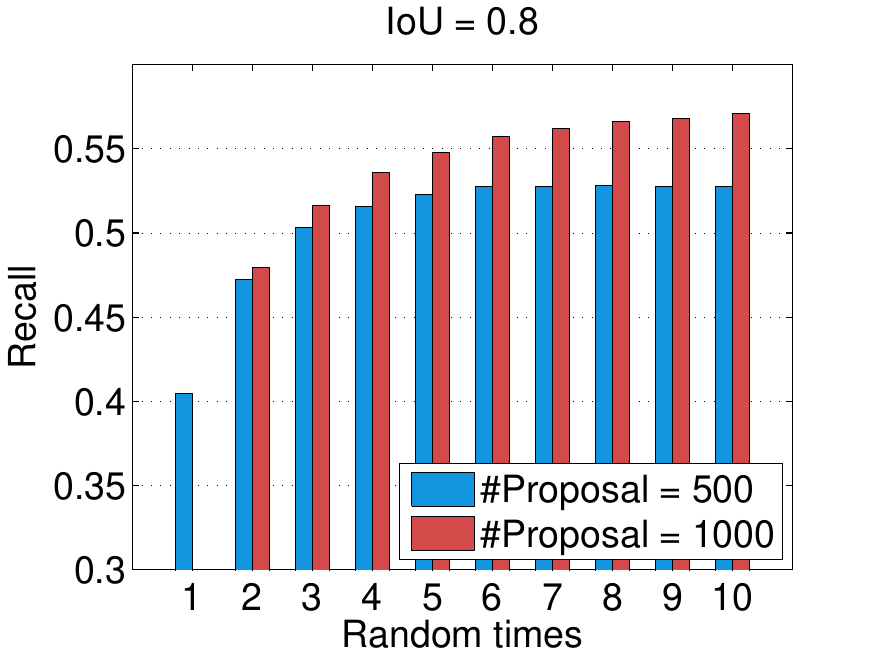}}
\caption{Comparison of recall rates with different randomized times on the PASCAL VOC 2007 test set. We report the results of both top 500 and 1000 proposals for a fair comparison. Note that the number of proposals generated by one-time randomized merging is less than 1000, so we cannot provide this result here. Best viewed in color.}
\label{fig:random_times}
\end{figure*}

\begin{figure*}[!t]
\centering
\subfigure[]{
\label{fig:stability_500} 
\includegraphics[width=0.32\linewidth]{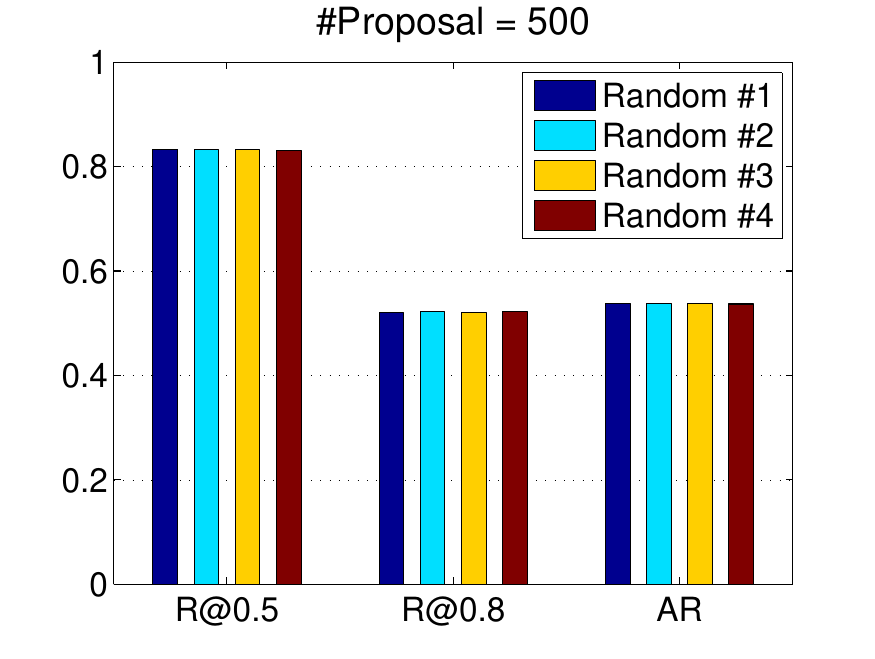}}
\subfigure[]{
\label{fig:stability_1000} 
\includegraphics[width=0.32\linewidth]{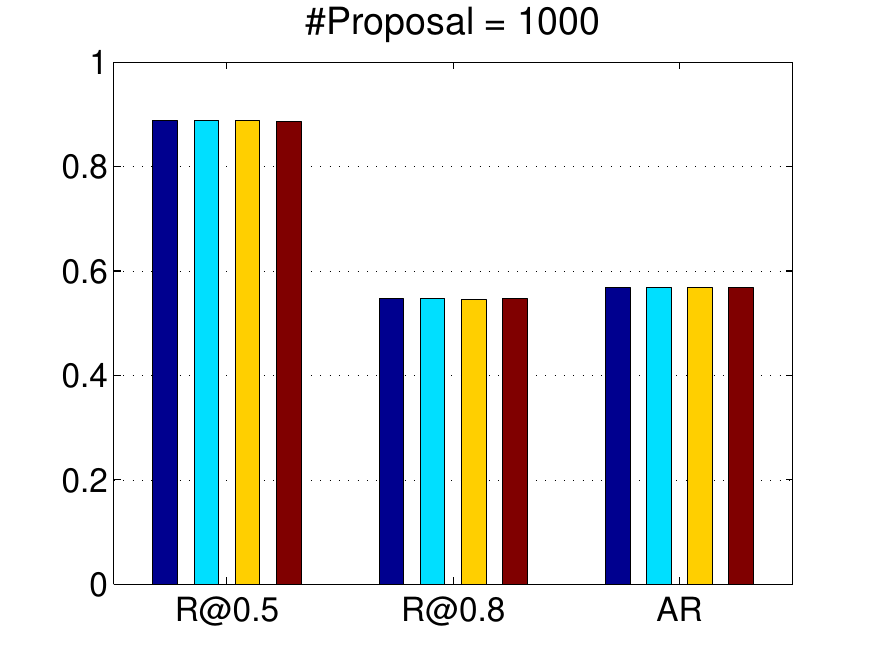}}
\subfigure[]{
\label{fig:stability_1462} 
\includegraphics[width=0.32\linewidth]{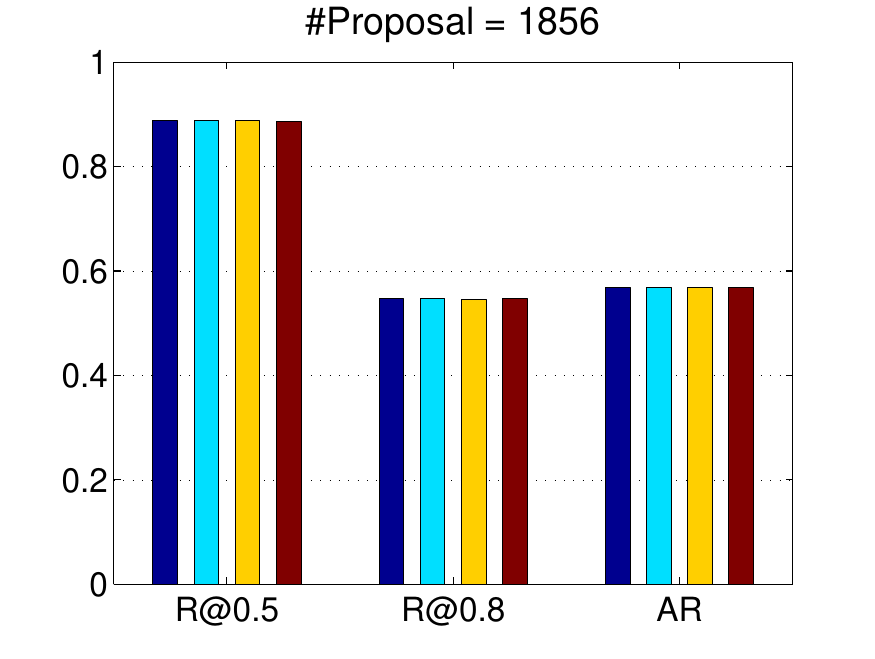}}
\caption{Comparison of recall rates in four groups of randomized merging experiments on the PASCAL VOC 2007 test set. We report the results using top ranked 500, 1000 and all of the proposals. R@0.5 and R@0.8 mean the recall rates with an IoU threshold of 0.5 and 0.8, respectively, and AR is the average recall. Best viewed in color.}
\label{fig:stability}
\end{figure*}

Typically, an IoU threshold of 0.5 is used to measure whether the target object is detected successfully in object detection tasks. However, as suggested in recent works ~\cite{hosang2015makes,zitnick2014edge, chen2015improving}, the proposals with an IoU of 0.5 cannot fit the ground truth objects well, usually resulting in a failure of subsequent object detectors. This reveals the fact that the recall rate with an IoU threshold of 0.5 is weakly correlated with the real detection performance. Hence, we also present the curves of recall rate with respect to the number of proposals at a more strict IoU threshold of 0.8, shown in Figure \ref{fig:VOC2007_result1}(e), to demonstrate the superiority of our method. We believe that our method may be more suitable for object detection systems owing to better localization accuracy and efficiency.
Besides, we compare the average recall (AR), considered to have a strong correlation with detection performance, as another important metric for evaluation. As shown in Figure \ref{fig:VOC2007_result1}(f), our method outperforms other state-of-the-art algorithms, which suggests that it is likely to achieve a better detection performance with the proposals generated by our method.

We also compare the performance on the PASCAL VOC 2012 validation set, as depicted in Figure \ref{fig:VOC2012_result1}. Note that RPN is trained with the data from both VOC 2007 and 2012 datasets, but RP and our method are learned on the VOC 2007 dataset without re-training here. Even though VOC 2012 is more challenging and larger in size, our method still achieves best performance over other state-of-the-art algorithms, again demonstrating the effectiveness of the proposed method. It is also shown that more improvement over other methods on the VOC 2012 is achieved than that on the VOC 2007.

We present some qualitative examples in Figure \ref{fig:visualization_results2}, including some random samples (top four rows) that contains two or more objects and some samples (the last row) that challenges our method. We find that, in most cases, our results match well with the ground-truth, and preserve the accurate object boundaries. The missed object are in part tiny ones, e.g., the distant and severely-occluded ones.

Figure \ref{fig:ar_area} analyzes the AR with regard to the ground truth objects in different sizes. It is shown that our method performs slightly worse than RPN if we only consider small-sized objects whose areas are less than 5k pixels. Nevertheless, our method yields best performance over other competitors in general, especially for recalling objects in larger size. Other grouping-based methods such as SS and CADM show similar results. One possible reason is that grouping-based methods depends heavily on the over-segmentations. In this case, the boundaries of small-sized objects are generally difficult to be well preserved, if the segmentation is not accurate enough. But this problem will not have a significant impact on larger-sized objects. Therefore, region-grouping-based approaches usually exhibit desirable ability to recall objects in relatively large size, but may fail to recall small-sized ones.

\subsection{Object Detection Performance}
Since most state-of-the-art object detectors rely on object proposals as a first preprocessing step, it is essential to evaluate the final detection performance with proposals generated by different methods. In this subsection, we conduct experiments to analyze the quality of proposals for object detection tasks. To this end, we use the Fast R-CNN detection framework ~\cite{girshick2015fast} using both CaffeNet ~\cite{krizhevsky2012imagenet} and VGG-Net \cite{simonyan2014very} as the benchmarks. The detectors are trained using PASCAL VOC 2007 trainval set, and tested using the test set for all the experiments here. We compare EB, SS, MCG2015 and RPN with the proposed method. For a fair comparison, we select top-1000 proposals for all of the methods in both training and testing stages. The mean average precision (mAP) and average precision (AP) for each of the 20 categories are shown in Table \ref{table:det}. 
It can be seen that our proposed method achieves the best mAPs of 58.6\% and 69.0\% using the Fast R-CNN with CaffeNet and VGG-Net, respectively, outperforming other state-of-the-art methods. This also verifies the effectiveness of our method for detection tasks.

\subsection{Generalization to Unseen Categories}
In addition, we conduct experiments on the ImageNet 2015 validation set to further evaluate the generalization ability to a wider scope of object categories. Note that all of the learning-based models (e.g., RP, RPN and ours) are trained on the PASCAL VOC dataset, and directly tested on the ImageNet 2015 validation set without re-training. The comparision of the experimental results are shown in Figure \ref{fig:ImageNet_result1}. It can be seen that our method has comparable performance with MCG2015, and surpasses other methods. No obvious deterioration in performance is observed on the ImageNet 2015 validation set, suggesting that our method does not exclusively fit the 20 specific categories of objects from the PASCAL VOC. In other words, our method is capable of capturing generic objectness information and generalizing to unseen categories. In addition, most state-of-the-art methods achieve similar performance to those on the PASCAL VOC, while RPN suffers from severe performance drop. One possible reason is that category information is exploited to learn class-specific detectors, which makes the RPN overfit 20 categories of objects from the PASCAL VOC.

\subsection{Evaluation of Randomized Merging Algorithm}
\label{subsec:Randomized_Inference}

\begin{table}[htp]
\centering
\begin{tabular}{c|c|c|c}
\hline
  & R@0.5 & R@0.8 & AR  \\
\hline
  Greedy &0.872 & 0.423 & 0.489  \\
\hline
  Random & 0.870 & 0.405 & 0.478  \\
\hline
\end{tabular}
\vspace{1pt}
\caption{Comparison of greedy merging and randomized merging on the PASCAL VOC 2007 test set. We report the results using top ranked 500 proposals. R@0.5 and R@0.8 indicate the recall rates with an IoU threshold of 0.5 and 0.8, respectively, and AR is the average recall.}
\label{table:random_and_greedy}
\end{table}

In this subsection, we evaluate the contribution of the proposed randomized merging algorithm. We compare the performance of conventional greedy and the proposed randomized merging algorithms on the PASCAL VOC 2007. In the first setting, we allow the randomized merging algorithm to be performed only one time in each recursion step, and the experimental results are shown in Table \ref{table:random_and_greedy}. It can be seen that greedy merging and one-time randomized merging achieve comparable results. However, with the increase of randomized times, our merging algorithm generates more diverse proposals. Here we conducted the experiments and compare the results obtained with different randomized times. The number of object proposals are fixed as 500 and 1000, respectively. Figure \ref{fig:random_times} clearly shows that the AR rate improves as the random times increases, and then goes near saturation eventually. This approach provides a notable gain in recall rates compared to greedy merging strategy. It is also shown that the recall rate with an IoU threshold of 0.5 first keep fixed and then drops as the random times increases, while that with 0.8 boosts consistently. This suggests that the predicted objectness scores are not accurate enough with small IoU values with the ground truth bounding boxes.

Another critical issue is that whether stable performance can be achieved by our proposed method, since we introduce randomness to the merging process. To better clarify this problem, we conduct four groups of experiments, and we repeat the randomized merging process for five times for each group. As shown in Figure \ref{fig:stability}, our method also exhibits great stability in recall rates and average recall with different numbers of object proposals.

\subsection{Efficiency Analysis}
In this subsection, we present the comparison of the efficiency of our model and the state-of-the-art methods. The execution time of MHS \cite{wang2015object} are directly taken from \cite{wang2015object} as its codes are not available. The deep-learning-based methods (i.e., ours and RPN) are conducted on a single NVIDIA TITAN X GPU, and the rest are carried out on a desktop with an Intel i7 3.4GHz CPU and 16G RAM. The average running time of all the methods for generating 1,000 proposals on the PASCAL VOC 2007 dataset are reported in Table \ref{table:efficiency}. It can be seen that window scoring methods achieve relatively high computational efficiency because of using very simple features and efficient scoring methods. Among those region grouping methods,  RP,  GOP and LPR run slightly faster than our method, but their performance are much inferior than ours on both PASCAL VOC and ImageNet datasets (see Figure \ref{fig:VOC2007_result1}, \ref{fig:VOC2012_result1} and \ref{fig:ImageNet_result1}). MCG2015 achieves comparable results, but it is extremely time-consuming. It is noteworthy that our method achieves the best performance among all methods while sharing quite high running efficiency. Specifically, for our method, it takes about 0.2s for feature extraction, and about 0.5s for one-time random merging. The random merging process is repeated for 8 times, thus the running time is 4.2s per image.

\begin{table}[htp]
\centering
\begin{tabular}{c|c|c|c}
\hline
  Type &  Method & Time & AR (\%)  \\
  \hline
  \multirow{3}{*}{Windows scoring} &
   Bing  ~\cite{cheng2014bing}  & 0.2 &  27.38\\
   & RPN ~\cite{ren2015faster} & 0.2 &   48.19\\
   & EB ~\cite{zitnick2014edge}  & 0.3 & 50.30\\
     \hline
    \multirow{8}{*}{Regions grouping} &
  RP ~\cite{manen2013prime} &1.0 & 46.30   \\
  & GOP ~\cite{krahenbuhl2014geodesic} &   1.0 & 49.38   \\
  & LPO ~\cite{krahenbuhl2015learning} &  1.1 & 50.98   \\
  & MHS ~\cite{wang2015object} &2.8 & 52.08  \\
  & SS  ~\cite{uijlings2013selective} &10.0 & 51.91   \\
  &  CACD ~\cite{xiao2015complexity} &22.0 & 52.30   \\
  & MCG2015 ~\cite{pont2015multiscale} & 30.0 &  57.45  \\
  & Ours & 4.2 &   \textbf{57.60}\\
\hline
\end{tabular}
\vspace{1pt}
\caption{Comparison of the average running time (second per image) for generating 1,000 proposals and the average recall (AR) on the PASCAL VOC 2007 test set.}
\label{table:efficiency}
\end{table}


\section{Conclusion}
\label{sec:conclusion}
In this paper, we have presented a simple yet effective approach to hierarchically segment object proposals by develping a deep architecture of recursive neural networks. We incorporate the similarity metric learning into the bottom-up region merging process for end-to-end training, rather than manually designing various representations. In addition, we introduce randomness into the greedy search to cope with the ambiguity in the process of merging regions, making the inference more robust against noise. Extensive experiments on standard benchmarks demonstrate the superiority of our approach over state-of-the-art approaches. In addition, the effectiveness of our method for real detection systems is also verified. In future work, this proposed framework can be tightly combined with category-specific object detection methods.

\section*{Acknowledgement}
This work was supported by State Key Development Program under Grant 2016YFB1001004, the National Natural Science Foundation of China under Grant 61622214, the Science and Technology Planning Project of Guangdong Province under Grant 2017A020208041, Special Program of the NSFC-Guangdong Joint Fund for Applied Research on Super Computation (the second phase), and Guangdong Natural Science Foundation Project for Research Teams under Grant 2017A030312006.

{
\bibliographystyle{IEEEtran}
\bibliography{reference}

\begin{thebibliography}{10}
\providecommand{\url}[1]{#1}
\csname url@samestyle\endcsname
\providecommand{\newblock}{\relax}
\providecommand{\bibinfo}[2]{#2}
\providecommand{\BIBentrySTDinterwordspacing}{\spaceskip=0pt\relax}
\providecommand{\BIBentryALTinterwordstretchfactor}{4}
\providecommand{\BIBentryALTinterwordspacing}{\spaceskip=\fontdimen2\font plus
\BIBentryALTinterwordstretchfactor\fontdimen3\font minus
  \fontdimen4\font\relax}
\providecommand{\BIBforeignlanguage}[2]{{%
\expandafter\ifx\csname l@#1\endcsname\relax
\typeout{** WARNING: IEEEtran.bst: No hyphenation pattern has been}%
\typeout{** loaded for the language `#1'. Using the pattern for}%
\typeout{** the default language instead.}%
\else
\language=\csname l@#1\endcsname
\fi
#2}}
\providecommand{\BIBdecl}{\relax}
\BIBdecl

\bibitem{girshick2014rich}
R.~Girshick, J.~Donahue, T.~Darrell, and J.~Malik, ``Rich feature hierarchies
  for accurate object detection and semantic segmentation,'' in \emph{Proc.
  IEEE Conf. Comput. Vis. Pattern Recognit.}\hskip 1em plus 0.5em minus
  0.4em\relax IEEE, Jun. 2014, pp. 580--587.

\bibitem{he2014spatial}
K.~He, X.~Zhang, S.~Ren, and J.~Sun, ``Spatial pyramid pooling in deep
  convolutional networks for visual recognition,'' in \emph{Proc. Eur. Conf.
  Comput. Vis.}, Sep. 2014, pp. 346--361.

\bibitem{wei2016hcp}
Y.~Wei, W.~Xia, M.~Lin, J.~Huang, B.~Ni, J.~Dong, Y.~Zhao, and S.~Yan, ``Hcp: A
  flexible cnn framework for multi-label image classification,'' \emph{IEEE
  transactions on pattern analysis and machine intelligence}, vol.~38, no.~9,
  pp. 1901--1907, 2016.

\bibitem{wang2017multi}
Z.~Wang, T.~Chen, G.~Li, R.~Xu, and L.~Lin, ``Multi-label image recognition by
  recurrently discovering attentional regions,'' in \emph{IEEE International
  Conference on Computer Vision}.\hskip 1em plus 0.5em minus 0.4em\relax IEEE,
  2017, pp. 464--472.

\bibitem{chen2018recurrent}
T.~Chen, Z.~Wang, G.~Li, and L.~Lin, ``Recurrent attentional reinforcement
  learning for multi-label image recognition,'' in \emph{Proc. of AAAI
  Conference on Artificial Intelligence}, 2018, pp. 6730--6737.

\bibitem{lee2011learning}
Y.~J. Lee and K.~Grauman, ``Learning the easy things first: Self-paced visual
  category discovery,'' in \emph{Proc. IEEE Conf. Comput. Vis. Pattern
  Recognit.}\hskip 1em plus 0.5em minus 0.4em\relax IEEE, Jun. 2011, pp.
  1721--1728.

\bibitem{cho2015unsupervised}
M.~Cho, S.~Kwak, C.~Schmid, and J.~Ponce, ``Unsupervised object discovery and
  localization in the wild: Part-based matching with bottom-up region
  proposals,'' \emph{arXiv preprint arXiv:1501.06170}, 2015.

\bibitem{uijlings2013selective}
J.~R. Uijlings, K.~E. van~de Sande, T.~Gevers, and A.~W. Smeulders, ``Selective
  search for object recognition,'' \emph{Int. J. Comput. Vis.}, vol. 104,
  no.~2, pp. 154--171, 2013.

\bibitem{manen2013prime}
S.~Manen, M.~Guillaumin, and L.~Van~Gool, ``Prime object proposals with
  randomized prim's algorithm,'' in \emph{Proc. IEEE Int. Conf. Comput.
  Vis.}\hskip 1em plus 0.5em minus 0.4em\relax IEEE, Dec. 2013, pp. 2536--2543.

\bibitem{xiao2015complexity}
Y.~Xiao, C.~Lu, E.~Tsougenis, Y.~Lu, and C.-K. Tang, ``Complexity-adaptive
  distance metric for object proposals generation,'' in \emph{Proc. IEEE Conf.
  Comput. Vis. Pattern Recognit.}\hskip 1em plus 0.5em minus 0.4em\relax IEEE,
  Jun. 2015, pp. 778--786.

\bibitem{socher2011parsing}
R.~Socher, C.~C. Lin, C.~Manning, and A.~Y. Ng, ``Parsing natural scenes and
  natural language with recursive neural networks,'' in \emph{Proc. Int. Conf.
  Mach. Learn.}\hskip 1em plus 0.5em minus 0.4em\relax ACM, Jun. 2011, pp.
  129--136.

\bibitem{ren2015faster}
S.~Ren, K.~He, R.~Girshick, and J.~Sun, ``Faster {R-CNN}: Towards real-time
  object detection with region proposal networks,'' \emph{arXiv preprint
  arXiv:1506.01497}, 2015.

\bibitem{zitnick2014edge}
C.~L. Zitnick and P.~Doll{\'a}r, ``Edge boxes: Locating object proposals from
  edges,'' in \emph{Proc. Eur. Conf. Comput. Vis.}\hskip 1em plus 0.5em minus
  0.4em\relax Springer, Sep. 2014, pp. 391--405.

\bibitem{cheng2014bing}
M.-M. Cheng, Z.~Zhang, W.-Y. Lin, and P.~Torr, ``{BING}: Binarized normed
  gradients for objectness estimation at 300fps,'' in \emph{Proc. IEEE Conf.
  Comput. Vis. Pattern Recognit.}\hskip 1em plus 0.5em minus 0.4em\relax IEEE,
  Jun. 2014, pp. 3286--3293.

\bibitem{alexe2010object}
B.~Alexe, T.~Deselaers, and V.~Ferrari, ``What is an object?'' in \emph{Proc.
  IEEE Conf. Comput. Vis. Pattern Recognit.}\hskip 1em plus 0.5em minus
  0.4em\relax IEEE, Jun. 2010, pp. 73--80.

\bibitem{alexe2012measuring}
B.~{Alexe}, T.~Deselaers, and V.~Ferrari, ``Measuring the objectness of image
  windows,'' \emph{IEEE Trans. Pattern Anal. Mach. Intell.}, vol.~34, no.~11,
  pp. 2189--2202, 2012.

\bibitem{zhang2011proposal}
Z.~Zhang, J.~Warrell, and P.~H. Torr, ``Proposal generation for object
  detection using cascaded ranking svms,'' in \emph{IEEE Conference on Computer
  Vision and Pattern Recognition (CVPR)}.\hskip 1em plus 0.5em minus
  0.4em\relax IEEE, 2011, pp. 1497--1504.

\bibitem{long2014fully}
J.~Long, E.~Shelhamer, and T.~Darrell, ``Fully convolutional networks for
  semantic segmentation,'' \emph{arXiv preprint arXiv:1411.4038}, 2014.

\bibitem{krahenbuhl2014geodesic}
P.~Kr{\"a}henb{\"u}hl and V.~Koltun, ``Geodesic object proposals,'' in
  \emph{Proc. Eur. Conf. Comput. Vis.}\hskip 1em plus 0.5em minus 0.4em\relax
  Springer, Sep. 2014, pp. 725--739.

\bibitem{carreira2012cpmc}
J.~Carreira and C.~Sminchisescu, ``{CPMC}: Automatic object segmentation using
  constrained parametric min-cuts,'' \emph{IEEE Trans. Pattern Anal. Mach.
  Intell.}, vol.~34, no.~7, pp. 1312--1328, 2012.

\bibitem{arbelaez2014multiscale}
P.~Arbelaez, J.~Pont-Tuset, J.~Barron, F.~Marques, and J.~Malik, ``Multiscale
  combinatorial grouping,'' in \emph{Proc. IEEE Conf. Comput. Vis. Pattern
  Recognit.}\hskip 1em plus 0.5em minus 0.4em\relax IEEE, Jun. 2014, pp.
  328--335.

\bibitem{rantalankila2014generating}
P.~Rantalankila, J.~Kannala, and E.~Rahtu, ``Generating object segmentation
  proposals using global and local search,'' in \emph{Proc. IEEE Conf. Comput.
  Vis. Pattern Recognit.}\hskip 1em plus 0.5em minus 0.4em\relax IEEE, Jun.
  2014, pp. 2417--2424.

\bibitem{bergh2013online}
M.~Bergh, G.~Roig, X.~Boix, S.~Manen, and L.~Gool, ``Online video seeds for
  temporal window objectness,'' in \emph{Proc. IEEE Int. Conf. Comput.
  Vis.}\hskip 1em plus 0.5em minus 0.4em\relax IEEE, Dec. 2013, pp. 377--384.

\bibitem{lin2017cross}
L.~Lin, G.~Wang, W.~Zuo, X.~Feng, and L.~Zhang, ``Cross-domain visual matching
  via generalized similarity measure and feature learning,'' \emph{IEEE
  transactions on pattern analysis and machine intelligence}, vol.~39, no.~6,
  pp. 1089--1102, 2017.

\bibitem{chen2016deep}
S.-Z. Chen, C.-C. Guo, and J.-H. Lai, ``Deep ranking for person
  re-identification via joint representation learning,'' \emph{IEEE
  Transactions on Image Processing}, vol.~25, no.~5, pp. 2353--2367, 2016.

\bibitem{wang2015object}
C.~Wang, L.~Zhao, S.~Liang, L.~Zhang, J.~Jia, and Y.~Wei, ``Object proposal by
  multi-branch hierarchical segmentation,'' in \emph{Proc. IEEE Conf. Comput.
  Vis. Pattern Recognit.}\hskip 1em plus 0.5em minus 0.4em\relax IEEE, June
  2015, pp. 3873--3881.

\bibitem{krahenbuhl2015learning}
P.~Kr{\"a}henb{\"u}hl and V.~Koltun, ``Learning to propose objects,'' in
  \emph{Proc. IEEE Conf. Comput. Vis. Pattern Recognit.}\hskip 1em plus 0.5em
  minus 0.4em\relax IEEE, Jun. 2015, pp. 1574--1582.

\bibitem{felzenszwalb2004efficient}
P.~F. Felzenszwalb and D.~P. Huttenlocher, ``Efficient graph-based image
  segmentation,'' \emph{Int. J. Comput. Vis.}, vol.~59, no.~2, pp. 167--181,
  2004.

\bibitem{girshick2015fast}
R.~Girshick, ``Fast {R-CNN},'' in \emph{Proc. IEEE Conf. Comput. Vis. Pattern
  Recognit.}\hskip 1em plus 0.5em minus 0.4em\relax IEEE, Jun. 2015, pp.
  1440--1448.

\bibitem{simonyan2014very}
K.~Simonyan and A.~Zisserman, ``Very deep convolutional networks for
  large-scale image recognition,'' \emph{arXiv preprint arXiv:1409.1556}, 2014.

\bibitem{chen2016disc}
T.~Chen, L.~Lin, L.~Liu, X.~Luo, and X.~Li, ``{DISC}: Deep image saliency
  computing via progressive representation learning.'' \emph{IEEE Trans. Neural
  Netw. Learn. Syst.}, 2016.

\bibitem{lin2016deep}
L.~Lin, K.~Wang, W.~Zuo, M.~Wang, J.~Luo, and L.~Zhang, ``A deep structured
  model with radius--margin bound for 3d human activity recognition,''
  \emph{International Journal of Computer Vision}, vol. 118, no.~2, pp.
  256--273, 2016.

\bibitem{lin2018active}
L.~Lin, K.~Wang, D.~Meng, W.~Zuo, and L.~Zhang, ``Active self-paced learning
  for cost-effective and progressive face identification,'' \emph{IEEE
  transactions on pattern analysis and machine intelligence}, vol.~40, no.~1,
  pp. 7--19, 2018.

\bibitem{chen2018knowledge}
T.~Chen, L.~Lin, R.~Chen, Y.~Wu, and X.~Luo, ``Knowledge-embedded
  representation learning for fine-grained image recognition,'' in \emph{Proc.
  of International Joint Conference on Artificial Intelligence}, 2018, pp.
  627--634.

\bibitem{wang2018deep}
Z.~Wang, T.~Chen, J.~Ren, W.~Yu, H.~Cheng, and L.~Lin, ``Deep reasoning with
  knowledge graph for social relationship understanding,'' in \emph{Proc. of
  International Joint Conference on Artificial Intelligence}, 2018, pp.
  2021--2018.

\bibitem{taskar2004max}
B.~Taskar, D.~Klein, M.~Collins, D.~Koller, and C.~D. Manning, ``Max-margin
  parsing.'' in \emph{EMNLP}.\hskip 1em plus 0.5em minus 0.4em\relax Citeseer,
  2004, p.~3.

\bibitem{bottou2012stochastic}
L.~Bottou, ``Stochastic gradient descent tricks,'' in \emph{Neural Networks:
  Tricks of the Trade}.\hskip 1em plus 0.5em minus 0.4em\relax Springer, 2012,
  pp. 421--436.

\bibitem{everingham2010pascal}
M.~Everingham, L.~Van~Gool, C.~K. Williams, J.~Winn, and A.~Zisserman, ``The
  pascal visual object classes (voc) challenge,'' \emph{Int. J. Comput. Vis.},
  vol.~88, no.~2, pp. 303--338, 2010.

\bibitem{russakovsky2014imagenet}
O.~Russakovsky, J.~Deng, H.~Su, J.~Krause, S.~Satheesh, S.~Ma, Z.~Huang,
  A.~Karpathy, A.~Khosla, M.~Bernstein \emph{et~al.}, ``Image{N}et large scale
  visual recognition challenge,'' \emph{Int. J. Comput. Vis.}, pp. 1--42, 2014.

\bibitem{hosang2015makes}
J.~Hosang, R.~Benenson, P.~Doll{\'a}r, and B.~Schiele, ``What makes for
  effective detection proposals?'' \emph{arXiv preprint arXiv:1502.05082},
  2015.

\bibitem{chen2015improving}
X.~Chen, H.~Ma, X.~Wang, and Z.~Zhao, ``Improving object proposals with
  multi-thresholding straddling expansion,'' in \emph{Proc. IEEE Conf. Comput.
  Vis. Pattern Recognit.}\hskip 1em plus 0.5em minus 0.4em\relax IEEE, Jun.
  2015, pp. 2587--2595.

\bibitem{jia2014caffe}
Y.~Jia, E.~Shelhamer, J.~Donahue, S.~Karayev, J.~Long, R.~Girshick,
  S.~Guadarrama, and T.~Darrell, ``{Caffe}: Convolutional architecture for fast
  feature embedding,'' in \emph{Proc. ACM Multimedia}.\hskip 1em plus 0.5em
  minus 0.4em\relax ACM, Nov. 2014, pp. 675--678.

\bibitem{pont2015multiscale}
J.~Pont-Tuset, P.~Arbelaez, J.~T. Barron, F.~Marques, and J.~Malik,
  ``Multiscale combinatorial grouping for image segmentation and object
  proposal generation,'' \emph{arXiv preprint arXiv:1503.00848}, 2015.

\bibitem{krizhevsky2012imagenet}
A.~Krizhevsky, I.~Sutskever, and G.~E. Hinton, ``{ImageNet} classification with
  deep convolutional neural networks,'' in \emph{Proc. Adv. Neural Inf.
  Process. Syst.}, Dec. 2012, pp. 1097--1105.

\end{thebibliography}
}

\begin{IEEEbiography}[{\includegraphics[width=1in,height=1.25in,clip,keepaspectratio]{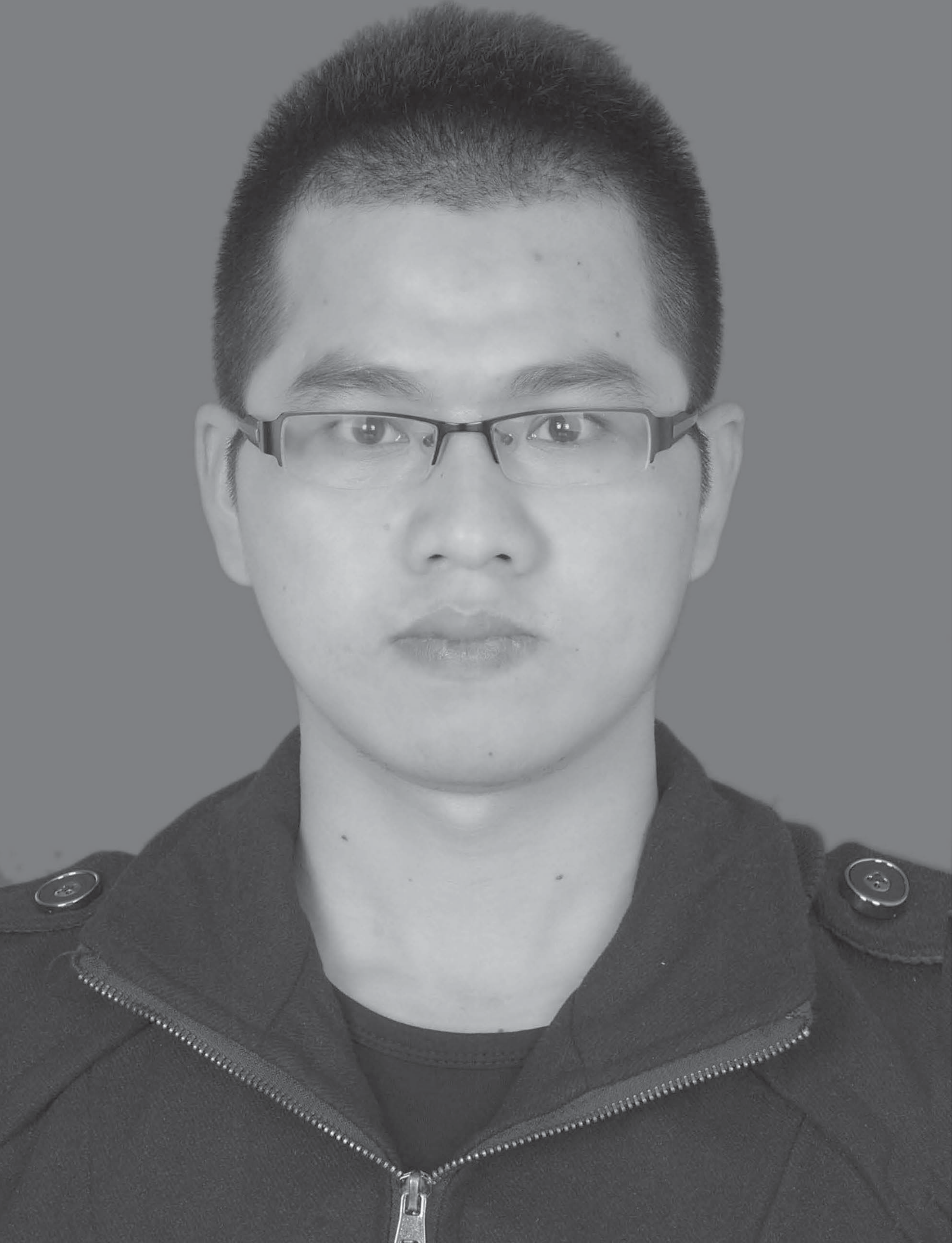}}]{Tianshui Chen} 
received the B.E. degree from School of Information and Science Technology, Sun Yat-sen University, Guangzhou, China, in 2013, where he is currently pursuing the Ph.D. degree in computer science with the School of Data and Computer Science. His current research interests include computer vision and machine learning.

\end{IEEEbiography}

\vspace{-25pt}

\begin{IEEEbiography}[{\includegraphics[width=1in,clip]{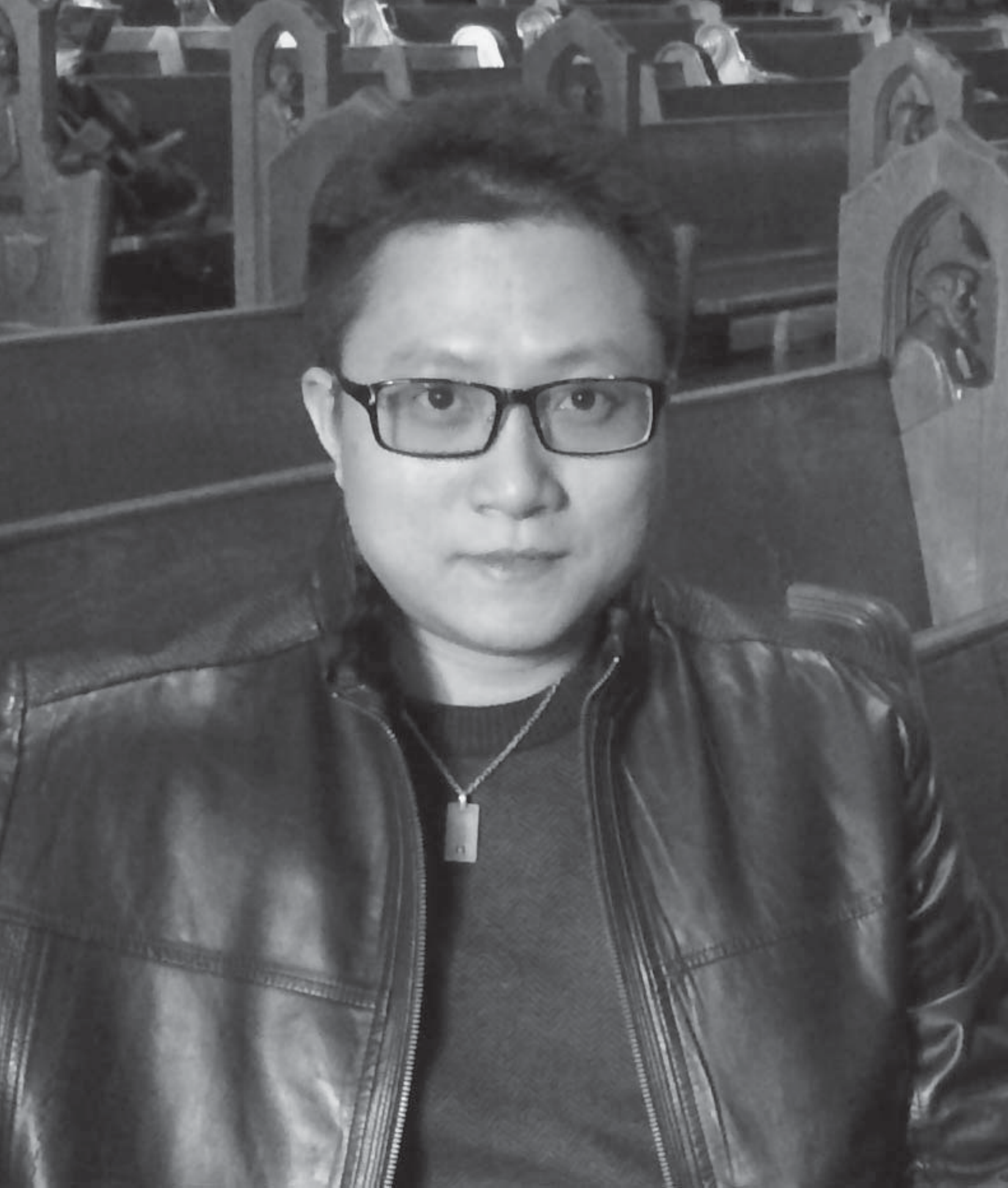}}]{Liang Lin} (M'09, SM'15) is the Executive R\&D Director of SenseTime Group Limited and a full Professor of Sun Yat-sen University. He is the Excellent Young Scientist of the National Natural Science Foundation of China. From 2008 to 2010, he was a Post-Doctoral Fellow at University of California, Los Angeles. From 2014 to 2015, as a senior visiting scholar he was with The Hong Kong Polytechnic University and The Chinese University of Hong Kong. He currently leads the SenseTime R\&D teams to develop cutting-edges and deliverable solutions on computer vision, data analysis and mining, and intelligent robotic systems. He has authorized and co-authorized on more than 100 papers in top-tier academic journals and conferences (e.g., 12 papers in TPAMI/IJCV and 50+ papers in CVPR/ICCV/NIPS/IJCAI). He has been serving as an associate editor of IEEE Trans. Human-Machine Systems, The Visual Computer and Neurocomputing. He served as Area/Session Chairs for numerous conferences such as ICME, ACCV, ICMR. He was the recipient of Best Paper Dimond Award in IEEE ICME 2017, Best Paper Runners-Up Award in ACM NPAR 2010, Google Faculty Award in 2012, Best Student Paper Award in IEEE ICME 2014, and Hong Kong Scholars Award in 2014. He is a Fellow of IET.
\end{IEEEbiography}

\vspace{-25pt}
\begin{IEEEbiography}[{\includegraphics[width=1in,height=1.25in,clip,keepaspectratio]{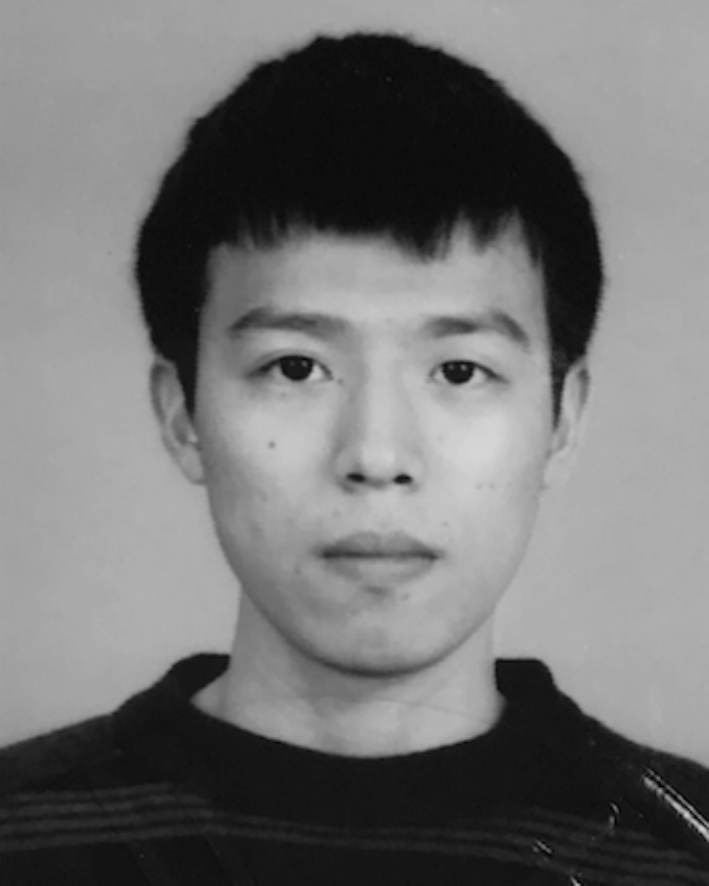}}]{Xian Wu} 
received the B.E. degree from the School of Software, Sun Yat-sen University, Guangzhou, China, in 2015, where he is currently pursuing the MS c. degree in Software Engineering with the School of Data and Computer Science. His current research interests include computer vision and machine learning.
\end{IEEEbiography}

\vspace{-25pt}
\begin{IEEEbiography}[{\includegraphics[width=1in,height=1.25in,clip,keepaspectratio]{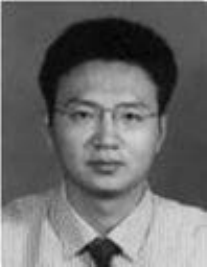}}]{Nong Xiao} received the BS and PhD degrees in computer science from the College of Computer at National University of Defense Technology (NUDT) in China, in 1990 and 1996, respectively. He is currently a professor in the State Key Laboratory of High Performance Computting at NUDT, China. His current research interests include large-scale storage system, network computing, and computer architecture. He has more than 130 publications to his credit in journals and international conferences including IEEE TSC, IEEE TMM, JPDC, JCST, HPCA, ICCAD, MIDDLEWARE, MSST, IPDPS, CLUSTER, SYSTOR and MASCOTS. He is a member of the IEEE and ACM.
\end{IEEEbiography}

\begin{IEEEbiography}[{\includegraphics[width=1in,height=1.25in,clip,keepaspectratio]{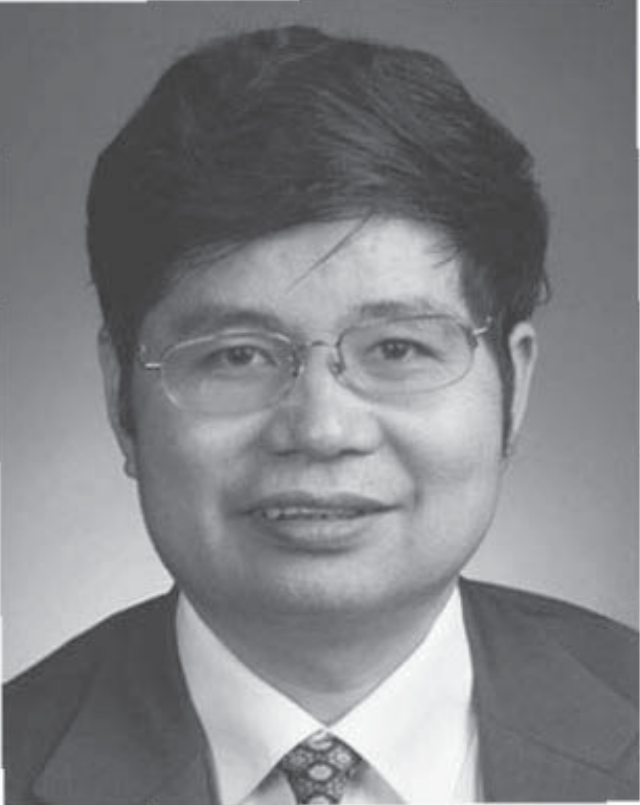}}]{Xiaonan Luo} was the Director of the National Engineering Research Center of Digital Life, Sun Yat-sen University, Guangzhou, China. He is currently a Professor with the School of Computer Science and Information Security, Guilin University of Electronic Technology, Guilin, China. He received the National Science Fund for Distinguished Young Scholars granted by the National Nature Science Foundation of China. His research interests include computer graphics, CAD, image processing, and mobile computing.
\end{IEEEbiography}

\end{document}